\documentclass{article} 
\usepackage{iclr2025_conference,times}

\usepackage{amsmath,amsfonts,bm}

\def\eqref#1{equation~\ref{#1}}

\def\1{\bm{1}}

\DeclareMathAlphabet{\mathsfit}{\encodingdefault}{\sfdefault}{m}{sl}
\SetMathAlphabet{\mathsfit}{bold}{\encodingdefault}{\sfdefault}{bx}{n}

\usepackage{hyperref}
\usepackage{url}

\usepackage{pifont}
\usepackage{float}
\usepackage{graphicx}
\usepackage{subcaption}
\usepackage{wrapfig}
\usepackage{amsmath}
\usepackage{amsfonts}
\usepackage{amsthm}

\usepackage{bm}
\usepackage{multirow}
\usepackage{multicol}
\usepackage{tabularx}
\usepackage{booktabs}
\usepackage{setspace}
\usepackage{xcolor,colortbl}
\usepackage{csquotes}
\usepackage{makecell}

\usepackage{algorithm}
\usepackage{algorithmicx}
\usepackage{algpseudocode}
\algrenewcommand\algorithmicrequire{\textbf{Input:}}
\algrenewcommand\algorithmicensure{\textbf{Output:}}

\usepackage{listings}
\usepackage[most,breakable]{tcolorbox}
\usepackage{fontawesome5}

\definecolor{Gold}{rgb}{1, 0.88, 0.22}
\definecolor{Silver}{rgb}{0.87, 0.87, 0.87}
\definecolor{Bronze}{rgb}{0.88, 0.62, 0.40}

\colorlet{GoldD}{Gold!95!black}
\colorlet{SilverD}{Silver!95!black}
\colorlet{BronzeD}{Bronze!95!black}
\newcommand{\mc}[2]{\textbf{\textcolor{#1D}{#2}}}

\newcommand{\medalbox}[2]{\begingroup\setlength{\fboxsep}{0.5pt}\colorbox{#1}{\strut #2}\endgroup}

\definecolor{pale}{RGB}{150 205 205}
\definecolor{blue}{RGB}{194, 213, 247}
\definecolor{orange}{RGB}{252, 225, 198}
\definecolor{green}{RGB}{155 205 155}
\definecolor{purple}{RGB}{146, 0, 199}
\definecolor{lightblue}{rgb}{0.92, 0.95, 1}
\definecolor{lightred}{rgb}{1, 0.9, 0.9}
\definecolor{deepblue}{rgb}{0, 0.4470, 0.7410}
\definecolor{deepyellow}{rgb}{0.9290, 0.6940, 0.1250}
\definecolor{deepgreen}{rgb}{0,0.5,0}
\usepackage{enumitem}
\setlist[itemize]{left=1em}

\title{PhysicsMinions: Winning Gold Medals in the Latest Physics Olympiads with a Coevolutionary Multimodal Multi-Agent System}

\author{
    Fangchen Yu$^{1,2}$\thanks{Equal contribution. $^{ \dagger}$Corresponding author.}~~, 
    Junchi Yao$^{1,4 \, *}$,
    Ziyi Wang$^{4}$,
    Haiyuan Wan$^{1,5}$,
    Youling Huang$^{1,6}$,\\ \textbf{
    Bo Zhang$^{1}$,
    Shuyue Hu$^{1}$,
    Dongzhan Zhou$^{1}$,
    Ning Ding$^{1,5}$,
    Ganqu Cui$^{1}$,} \\ \textbf{
    Lei Bai$^{1}$,
    Wanli Ouyang$^{1,3}$,
    Peng Ye$^{1,3 \dagger}$}\\
    $^1$Shanghai AI Laboratory,
    $^2$CUHK-Shenzhen,
    $^3$CUHK,
    $^4$UESTC,
    $^5$Tsinghua University,
    $^6$DUT
}

\iclrfinalcopy 
\begin{document}

\maketitle

\begin{abstract}

Physics is central to understanding and shaping the real world, and the ability to solve physics problems is a key indicator of real-world physical intelligence. Physics Olympiads, renowned as the crown of competitive physics, provide a rigorous testbed requiring complex reasoning and deep multimodal understanding, yet they remain largely underexplored in AI research. Existing approaches are predominantly single-model based, and open-source MLLMs rarely reach gold-medal-level performance. To address this gap, we propose \textsc{PhysicsMinions}, a coevolutionary multi-agent system for Physics Olympiad. Its architecture features three synergistic studios: a Visual Studio to interpret diagrams, a Logic Studio to formulate solutions, and a Review Studio to perform dual-stage verification. The system coevolves through an iterative refinement loop where feedback from the Review Studio continuously guides the Logic Studio, enabling the system to self-correct and converge towards the ground truth. Evaluated on the HiPhO benchmark spanning 7 latest physics Olympiads, \textsc{PhysicsMinions} delivers three major breakthroughs: \textbf{(i) Strong generalization:} it consistently improves both open-source and closed-source models of different sizes, delivering clear benefits over their single-model baselines; \textbf{(ii) Historic breakthroughs:} it elevates open-source models from only 1–2 to 6 gold medals across 7 Olympiads, achieving the first-ever open-source gold medal in the latest International Physics Olympiad (IPhO) under the average-score metric; and \textbf{(iii) Scaling to human expert:} it further advances the open-source Pass@32 score to 26.8/30 points on the latest IPhO, ranking 4$^\text{th}$ of 406 contestants and far surpassing the top single-model score of 22.7 (ranked 22$^\text{nd}$). Generally, \textsc{PhysicsMinions} offers a generalizable framework for Olympiad-level problem solving, with the potential to extend across disciplines.

\end{abstract}


\begin{figure}[H]
    \centering
    \vspace{-1mm}
    
    \includegraphics[width=.92\linewidth]{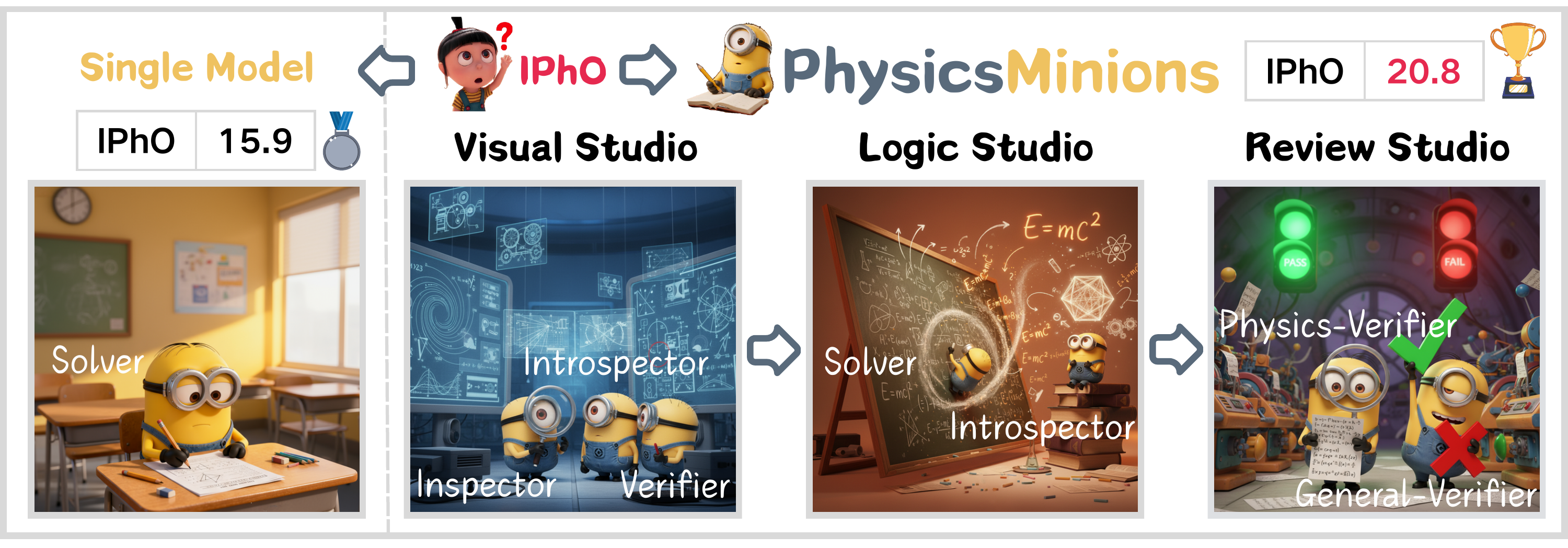}
    \vspace{-1mm}
    
    \caption{Illustration of \textsc{PhysicsMinions}, a coevolutionary multimodal multi-agent system comprising three studios: the \emph{Visual Studio} for visual extraction, the \emph{Logic Studio} for solution refinement, and the \emph{Review Studio} for dual verification. Like a single ``Minion,'' one agent is limited, but together they form a system that raises Intern-S1 from 15.9 (silver) to 20.8 (gold) in the latest IPhO.
    }
    \label{fig:overview}
\end{figure}

\newpage
\section{Introduction}

A deep understanding of physics is essential for shaping the real world, and the ability to solve physics problems is a critical step toward developing real-world physical intelligence~\citep{2025physics,2025physicsarena}. Physics Olympiads stand out as the crown of competitive problem solving, requiring both complex physics reasoning and advanced multimodal understanding. Yet results from the HiPhO benchmark~\citep{2025hipho}, dedicated to physics Olympiads, expose the limitations of current single-model paradigms. On the latest International Physics Olympiad (IPhO), only three closed-source models barely surpassed the gold medal threshold, while no open-source MLLM achieved gold, with most scoring near or below the bronze cutoff. These outcomes highlight both the formidable difficulty of Olympiad-level physics and the limitation of single-model approaches.

Encouragingly, advances in multi-agent systems have demonstrated the potential of agent-driven reasoning~\citep{madaan2023self,wang2025learning}. At the latest International Mathematical Olympiad (IMO), leading single models such as GPT-5 and Gemini-2.5-Pro scored well below the bronze medal line\footnote{See results at \url{https://matharena.ai/?comp=imo--imo_2025}.}, yet with a multi-agent framework they solved 5 of 6 problems, reaching gold-level performance~\citep{huang2025gemini}. This shows the promise of multi-agent paradigms in overcoming reasoning bottlenecks. However, unlike the purely text-based IMO, physics Olympiads present unique challenges:
\textbf{(1) Complex physical reasoning}, involving equation derivations, applications of laws and theorems, and long-horizon dependencies;
\textbf{(2) Multimodal understanding}, as figures, plots, and diagrams often contain indispensable information.
These challenges make existing mathematical multi-agent frameworks insufficient for direct application to physics reasoning.

To address this gap, we present \textbf{\textsc{PhysicsMinions}}, the pioneering coevolutionary multimodal multi-agent system tailored for physics Olympiads. As illustrated in Fig.~\ref{fig:overview}, each agent acts like a ``Minion'': a single one may be unable to solve Olympiad-level problems alone, but through coevolutionary collaboration, the system achieves complex reasoning. \textsc{PhysicsMinions} is organized into three specialized studios:
\textbf{(1) Visual Studio}, which transforms the visual inputs into structured information;
\textbf{(2) Logic Studio}, where a solver generates an initial solution and an introspector iteratively improves it;
\textbf{(3) Review Studio}, which employs both a physics-specific and a general verifier for dual-stage checking.
The three studios interact in a coevolutionary loop: Visual Studio validates and refines extracted information, Review Studio feeds verification results back to guide correction, and Logic Studio integrates these signals to iteratively refine solutions, enabling the system to substantially enhance multimodal physical reasoning and progressively approach the ground truth.

Evaluations on the HiPhO benchmark \citep{2025hipho}, spanning 7 latest physics Olympiads, validate the effectiveness of \textsc{PhysicsMinions} with three key breakthroughs:
\textbf{(1) Strong generalization:} the system consistently improves both closed- and open-source models with different scales, outperforming their single-model baselines;
\textbf{(2) Historic breakthroughs:} open-source models that achieved only 1-2 gold medals alone now obtain 6 in 7 Olympiads, including the first-ever gold in the latest IPhO;
\textbf{(3) Scaling to human expert:} the open-source Pass@32 score on IPhO reaches 26.8/30 points, placing 4$^\text{th}$ among 406 contestants and well above the top single-model result of 22.7 (22$^\text{nd}$ place).
These show that \textsc{PhysicsMinions} elevates MLLMs to gold-medal performance and toward human-expert levels, highlighting its potential as a general framework for problem solving.

Our work makes the following contributions:
\begin{itemize}
    \item \textbf{A new paradigm for Olympiad-level physics reasoning.}  
    We introduce \textsc{PhysicsMinions}, a pioneering coevolutionary multimodal multi-agent system tailored for physics Olympiads. Unlike the single-model paradigm, our system leverages coevolutionary agents and cross-modal information integration to push beyond the single-model ceiling on the latest physics Olympiads.

    \item \textbf{Coevolutionary framework design.}  
    We develop a coevolutionary framework with three specialized studios, where the \emph{Visual Studio} validates and refines visual information, the \emph{Review Studio} performs dual-stage verification to guide refinement in the \emph{Logic Studio}. All three interact in a coevolutionary loop, thereby strengthening multimodal physical reasoning.

    \item \textbf{Historic breakthroughs.}  
    Our framework consistently improves both closed- and open-source models, raising open-source results from 1-2 to 6 gold medals across 7 Olympiads and delivering the first-ever open-source gold in the latest IPhO. It further achieves 4$^\text{th}$ place among 406 contestants with Pass@32 score via open-source Intern-S1, surpassing 99\% contestants.
\end{itemize}

\section{Related Work}

\textbf{Challenges in Multimodal Physics Olympiads.}  
Benchmarks such as PhysUniBench~\citep{2025physunibench} and SeePhys~\citep{2025seephys} highlight the difficulty of combining physics reasoning with visual information in multimodal problems. OlympiadBench~\citep{2024olympiadbench} and OlympicArena~\citep{2024olympicarena} show the formidable complexity of physics Olympiad problems requiring long-horizon reasoning. Building on both, HiPhO~\citep{2025hipho} (High School Physics Olympiad Benchmark) poses dual challenges from multimodal understanding and Olympiad-level physics reasoning, where evaluations reveal that open-source MLLMs rarely achieve gold-medal performance, underscoring the difficulty of multimodal physical reasoning at the Olympiad level.

\textbf{Reasoning Methods and Multi-agent Frameworks.}  
Various methods have been developed to enhance the reasoning capabilities of (M)LLMs. Single-model approaches include Chain-of-Thought~\citep{wei2022chain} to strengthen step-by-step reasoning, Best-of-N~\citep{stiennon2020learning} to select the most reliable answer from multiple attempts, and Self-MoA~\citep{li2025rethinking} to aggregate diverse outputs through varied prompting. More recently, multi-agent frameworks such as Self-Refine~\citep{madaan2023self} and Generative Self-Refinement~\citep{wang2025learning} adopt self-reflection, enabling models to improve their own solutions. Mathematical multi-agent systems have also gained attention, exemplified by the framework proposed for IMO~\citep{huang2025gemini}, which applies self-verification and self-improvement to pure-text mathematical problems; however, such methods cannot be directly extended to the multimodal challenges of physics Olympiads.
\section{\textsc{PhysicsMinions}: A Coevolutionary Multi-Agent System}
\label{sec:system}

\begin{figure}[H]
    \centering
    \includegraphics[width=\textwidth]{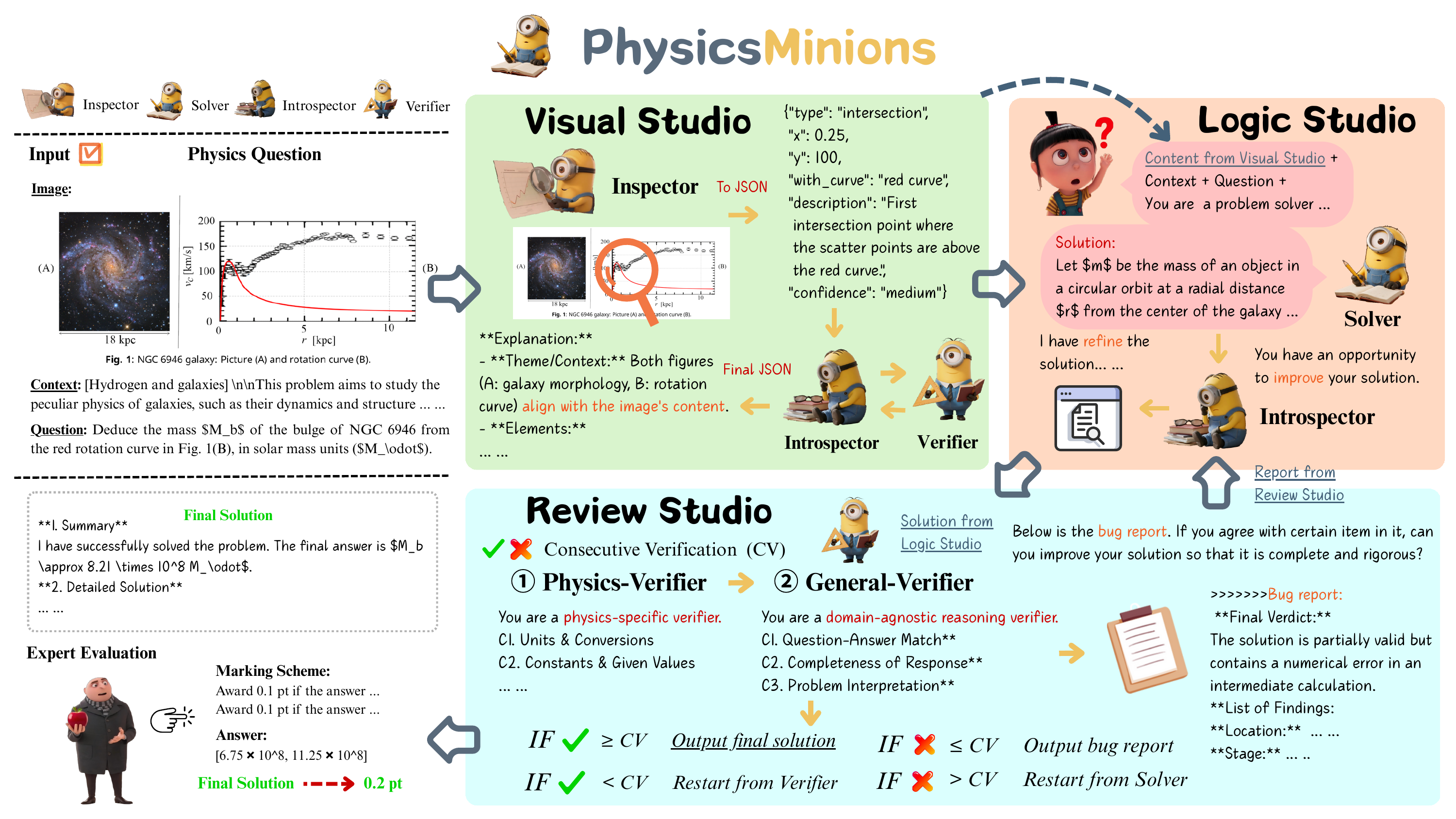}
    \caption{Overview of \textsc{PhysicsMinions}, a coevolutionary multimodal multi-agent system. Given a multimodal problem, the \emph{Visual Studio} extracts structured visual information. The \emph{Logic Studio} generates an initial solution and improves it. The \emph{Review Studio} then conducts dual-stage verification; failures trigger bug reports returned to the \emph{Logic Studio} for further revision. This loop continues until the solution passes consecutive checks, forming the coevolutionary process.}
    \label{fig:framework}
\end{figure}

\subsection{Overview of Framework}  

\textsc{PhysicsMinions} consists of three coevolutionary studios: the \emph{Visual Studio}, the \emph{Logic Studio}, and the \emph{Review Studio}. Given a multimodal problem with diagrams or plots, the Visual Studio first observes, validates, and reflects on the input to extract structured information, which is then passed to the Logic Studio. In the Logic Studio, a solver generates an initial solution and an introspector refines it through self-improvement before passing it on. The Review Studio then applies dual-stage verification: the Physics-Verifier checks physical consistency (e.g., constants and units), while the General-Verifier conducts more detailed inspections of logic, reasoning, and calculations. If either stage fails, a detailed bug report is returned to the Logic Studio, where the introspector revises the solution and resubmits it to Review Studio for verification. This process repeats until the solution passes a predefined number of consecutive verifications (CV), which is the only hyperparameter in the system. A solution that passes CV checks consecutively is accepted as the final solution; if it fails CV times consecutively, the solver regenerates a new candidate solution. This collaborative critique-and-refine cycle defines the system's coevolutionary process, with CV set to 2 by default (see Section~\ref{sec:ablation} for analysis). We next provide a detailed introduction of each studio.

\subsection{Visual Studio}
\label{framework:visual_studio}

\textbf{Pipeline.} The Visual Studio consists of three cooperative agents: an \emph{Inspector}, an \emph{Introspector}, and a \emph{Verifier}. Given a physics problem with multimodal diagrams, the Inspector first determines the image type (e.g., \texttt{plot}, \texttt{curve}, \texttt{free-body}, \texttt{circuit}, etc.) and then extracts task-relevant details. For a plot, it records axis labels, ranges, and tick values; for curves, it distinguishes colors and line styles, identifying features such as endpoints and peaks; for free-body diagrams, it lists objects together with the forces and their directions. These features are converted into a structured JSON description, which the Introspector refines to be self-contained, consistent, and faithful to the image. The Verifier then checks for errors such as wrong values, missing elements, or inconsistencies. If problems are found, a bug report is returned to the Introspector for revision before resubmission. Once the description passes a predefined number of consecutive verifications (CV), it is accepted as an accurate representation of the visual information. This structured output, rather than the raw image, is then passed to the Logic Studio along with the problem statement.  

\textbf{Innovation.}  
Unlike prior methods that feed raw images directly into the model, the Visual Studio converts multimodal inputs into structured JSON through iterative observation, introspection, and verification. This process reduces ambiguity, ensures consistency, and bridges visual perception with symbolic reasoning. As shown in our ablation studies (Section~\ref{sec:ablation}), such validated and refined representations yield significantly better performance than raw images, underscoring the importance of structured visual information for complex physics reasoning.

\subsection{Logic Studio}
\label{framework:logic_studio}

\textbf{Pipeline.}  
The Logic Studio consists of two cooperative agents: a \emph{Solver} and an \emph{Introspector}. Given the problem statement and the structured JSON from the Visual Studio, the Solver generates a \emph{structured solution} with two components: \textbf{(i) a Summary}, which declares the solution as \emph{Complete} or \emph{Partial}, provides the final answer if complete (or rigorously proven results if partial), and outlines a method sketch; and \textbf{(ii) a Detailed Solution}, which presents a step-by-step analysis with all equations written in \TeX. The Introspector then improves this solution, focusing on equation derivation, numerical calculations, and overall consistency. The revised solution is then passed to the Review Studio for dual-stage verification. If verification fails, the Introspector receives a bug report and revises the solution; when it disagrees with certain items, it provides explicit justifications to avoid repeated misunderstandings. This loop continues until the solution passes the required number of consecutive verifications (CV) or, after persistent failures, the Solver regenerates a new candidate.

\textbf{Innovation.}  
The Logic Studio enforces a \emph{structured solution format} (Summary + Detailed Solution) that makes reasoning explicit and errors traceable, enabling targeted bug reports and precise refinements. Combined with verification feedback, the Solver–Introspector collaboration forms a critique-and-refine coevolution: solutions are iteratively corrected where they fail, reinforced where they hold, and progressively driven toward the ground truth.

\subsection{Review Studio}
\label{framework:review_studio}

\textbf{Pipeline.}  
The Review Studio performs dual-stage verification with a \emph{Physics-Verifier} followed by a \emph{General-Verifier}. The Physics-Verifier first carries out domain-specific checks, including coarse validation of units and physical constants, as well as finer checks of assumptions and physical consistency (e.g., detecting when a formula is applied to the wrong type of quantity). If the solution fails, a bullet-point bug report is generated and returned directly to the Logic Studio for refinement, bypassing the second stage. If it passes, the solution proceeds to the General-Verifier, which begins with coarse checks for completeness and problem understanding (e.g., missing sub-questions or misinterpretations), followed by fine-grained step-by-step verification of logical consistency, reasoning flow, and algebraic or numerical calculations. Failures at this stage trigger a comprehensive bug report. This report is then returned to the Logic Studio, where the Introspector revises the solution until it passes both stages consecutively.

\textbf{Innovation.}  
The Review Studio introduces two key innovations. First, the dual-stage design separates physics-specific and general verification: the Physics-Verifier quickly filters out domain errors, while the General-Verifier ensures broader logical soundness across disciplines. Second, both verifiers adopt a coarse-to-fine strategy, starting with high-level checks before moving to detailed verification. This layered design improves error coverage, and the structured bug reports provide clear guidance for the Introspector, enabling efficient and precise corrections toward ground truth.
\begin{table}[t]
\centering
\caption{Evaluation results on 7 latest physics Olympiads from the HiPhO benchmark using the exam score metric. \medalbox{Gold!50}{Gold}, \medalbox{Silver!70}{Silver}, and \medalbox{Bronze!40}{Bronze} indicate scores above the medal thresholds, following HiPhO. Only the theoretical parts of the exams are considered, so Full Mark (Model) $\leq$ Full Mark (Human). Top-1 Score (Human) is the highest score among human medalists, while Top-1 Score (Model) is the best single-model score on the HiPhO leaderboard.}
\label{tab:main_results}
\small
\resizebox{\textwidth}{!}{%
\setlength{\tabcolsep}{1.8pt}
\begin{tabular}{lccccccc|ccc}
\toprule
\textbf{Latest Physics Olympiads} & \textbf{IPhO} & \textbf{APhO} & \textbf{EuPhO} & \textbf{NBPhO} & \textbf{PanPhO} & \textbf{PanMechanics} & \textbf{F=MA} & \multicolumn{3}{c}{\textbf{Medal}} \\ \midrule
Full Mark (Human) & 30.0 & 30.0 & 30.0 & 72.0 & 100.0 & 100.0 & 25.0 \\
Full Mark (Model) & 29.4 & 30.0 & 29.0 & 43.5 & 100.0 & 100.0 & 25.0 \\
Top-1 Score (Human) & 29.2 & 30.0 & 27.0 & 53.2 & 81.0 & 62.0 & 25.0 \\
Top-1 Score (Model) & 22.7 & 27.9 & 14.9 & 34.1 & 60.3 & 72.1 & 22.8 \\
\rowcolor{Gold!40}
Gold Medal   & 19.7 & 23.3 & 16.5 & 28.6 & 41.5 & 52.0 & 15.0 & \multicolumn{3}{>{\cellcolor{white}}l}{\textcolor{Gold}{\faMedal}}  \\
\rowcolor{Silver!60}
Silver Medal & 12.1 & 18.7 & 9.8 & 20.1 & 28.5 & 36.0 & 11.0 & \multicolumn{3}{>{\cellcolor{white}}c}{\textcolor{Silver}{\faMedal}}  \\
\rowcolor{Bronze!40}
Bronze Medal & 7.2 & 13.1 & 5.8 & 15.2 & 14.5 & 20.0 & 9.0 & \multicolumn{3}{>{\cellcolor{white}}r}{\textcolor{Bronze}{\faMedal}}  \\ \midrule
%
Gemini-2.5-Flash-Thinking & \cellcolor{Gold!35}20.2 & \cellcolor{Gold!35}27.4 & \cellcolor{Silver!60}13.2 & \cellcolor{Gold!35}29.0 & \cellcolor{Gold!35}44.6 & \cellcolor{Gold!35}60.5 & \cellcolor{Gold!35}17.8 & ~\mc{Gold}{6} & \mc{Silver}{1} & \mc{Bronze}{0}\\

\;\;\textbf{+ \textsc{PhysicsMinions}} & \cellcolor{Gold!35}21.5 & \cellcolor{Gold!35}28.0 & \cellcolor{Gold!35}16.5 & \cellcolor{Gold!35}33.3 & \cellcolor{Gold!35}57.8 & \cellcolor{Gold!35}72.0 & \cellcolor{Gold!35}24.0 & ~\mc{Gold}{7} & \mc{Silver}{0} & \mc{Bronze}{0}\\
\arrayrulecolor[gray]{0.7}\midrule\arrayrulecolor{black}

Intern-S1 & \cellcolor{Silver!60}15.9 & \cellcolor{Silver!60}21.7 & \cellcolor{Bronze!40}9.0 & \cellcolor{Silver!60}23.0 & \cellcolor{Silver!60}41.1 & \cellcolor{Gold!35}60.4 & \cellcolor{Gold!35}18.4 & ~\mc{Gold}{2} & \mc{Silver}{4} & \mc{Bronze}{1}\\

\;\;\textbf{+ \textsc{PhysicsMinions}} & \cellcolor{Gold!35}20.8 & \cellcolor{Gold!35}25.2 & \cellcolor{Silver!60}10.1 & \cellcolor{Gold!35}28.9 & \cellcolor{Gold!35}46.8 & \cellcolor{Gold!35}68.7 & \cellcolor{Gold!35}22.7 & ~\mc{Gold}{6} & \mc{Silver}{1} & \mc{Bronze}{0}\\
\arrayrulecolor[gray]{0.7}\midrule\arrayrulecolor{black}

InternVL3.5-241B-A28B & \cellcolor{Bronze!40}12.0 & \cellcolor{Silver!60}21.1 & \cellcolor{Bronze!40}9.4 & \cellcolor{Silver!60}22.6 & \cellcolor{Bronze!40}24.9 & \cellcolor{Gold!35}54.7 & \cellcolor{Silver!60}14.0 & ~\mc{Gold}{1} & \mc{Silver}{3} & \mc{Bronze}{3}\\

\;\;\textbf{+ \textsc{PhysicsMinions}} & \cellcolor{Gold!35}20.9 & \cellcolor{Gold!35}24.6 & \cellcolor{Silver!60}9.8 & \cellcolor{Gold!35}29.6 & \cellcolor{Gold!35}46.2 & \cellcolor{Gold!35}66.7 & \cellcolor{Gold!35}21.0 & ~\mc{Gold}{6} & \mc{Silver}{1} & \mc{Bronze}{0}\\
\arrayrulecolor[gray]{0.7}\midrule\arrayrulecolor{black}

Qwen2.5VL-32B-Instruct & \cellcolor{Bronze!40}9.9 & \cellcolor{Bronze!40}16.5 & \cellcolor{Bronze!40}6.9 & \cellcolor{Bronze!40}15.3 & \cellcolor{Bronze!40}22.5 & \cellcolor{Bronze!40}28.1 & 7.6 & ~\mc{Gold}{0} & \mc{Silver}{0} & \mc{Bronze}{6}\\

\;\;\textbf{+ \textsc{PhysicsMinions}} & \cellcolor{Silver!60}12.4 & \cellcolor{Bronze!40}17.7 & \cellcolor{Bronze!40}9.0 & \cellcolor{Silver!60}21.0 & \cellcolor{Silver!60}29.5 & \cellcolor{Silver!60}36.0 & \cellcolor{Silver!60}12.0 & ~\mc{Gold}{0} & \mc{Silver}{5} & \mc{Bronze}{2}\\
\bottomrule
\end{tabular}%
}
\end{table}

\section{Experiments}

\subsection{Experimental Setup}

\textbf{Evaluation.} We evaluate \textsc{PhysicsMinions} on seven latest physics Olympiads from the HiPhO benchmark~\citep{2025hipho}, which covers both international and regional competitions. Following the HiPhO setup, we fixed the model temperature at 0.6 and applied both answer-level and step-level evaluation based on the official marking schemes, enabling direct human-level comparison. For each exam, we record the average score across three repeated inference runs.

\textbf{Models.}  
We evaluate four representative MLLMs, spanning closed- and open-source models at different scales. \textbf{(1) Gemini-2.5-Flash-Thinking}~\citep{Gemini-2.5}, a closed-source model with strong reasoning ability, ranks 2$^\text{nd}$ overall on the HiPhO leaderboard. \textbf{(2) Intern-S1}~\citep{Intern-S1} is the top-ranked open-source MLLM on the HiPhO leaderboard. \textbf{(3) InternVL3.5-241B-A28B}~\citep{InternVL3.5}, a flagship open-source model, achieves state-of-the-art results across diverse benchmarks. \textbf{(4) Qwen2.5VL-32B-Instruct}~\citep{Qwen2.5-VL}, a medium-scale open-source model, allows us to assess how \textsc{PhysicsMinions} enhances models with more limited capacity.

\subsection{Overall Breakthroughs with \textsc{PhysicsMinions}}

The main results in Table~\ref{tab:main_results} show that \textsc{PhysicsMinions} achieves three major breakthroughs on 7 latest physics Olympiads, significantly advancing multimodal physical reasoning. 

\begin{figure}[H]
    \centering
    \includegraphics[width=1\linewidth]{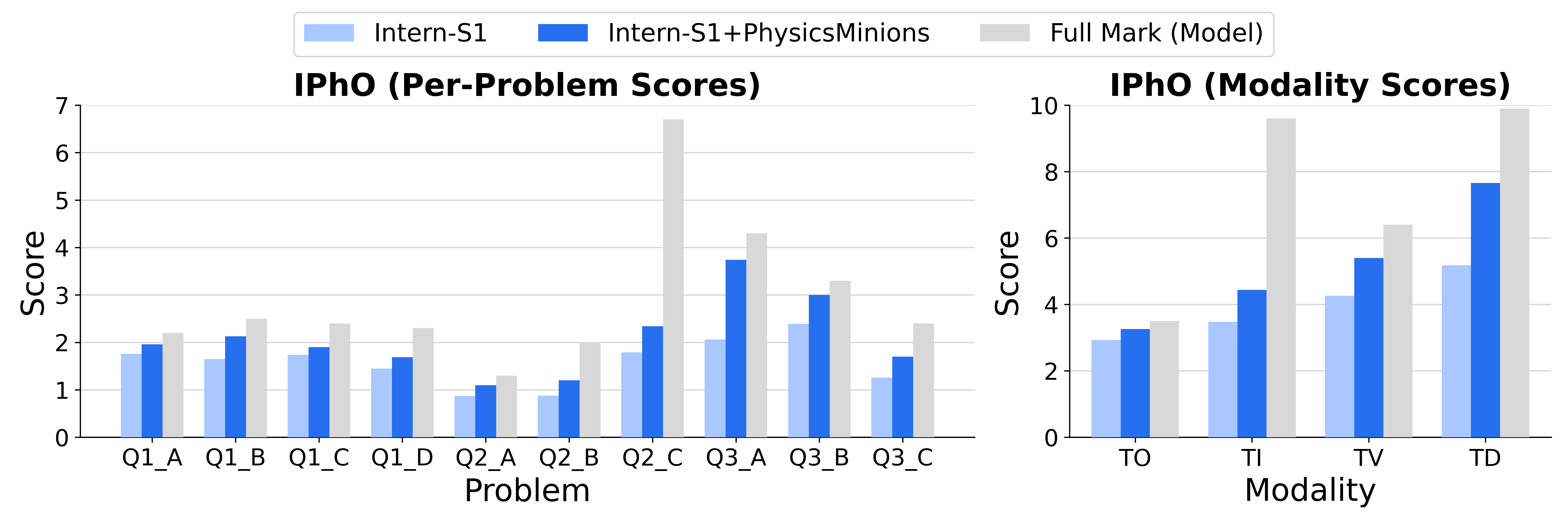}
    \caption{Performance improvement of Intern-S1 with \textsc{PhysicsMinions} on the latest IPhO, shown by per-problem and modality scores. The HiPhO benchmark defines four modality types: TO = Text-Only, TI = Text+Illustration Figure, TV = Text+Variable Figure, and TD = Text+Data Figure.}
    \label{fig:score_ipho_2025}
\end{figure}

\textbf{(1) \textsc{PhysicsMinions} delivers consistent improvements over single-model baselines.}  
The gains hold across both closed- and open-source models, regardless of architecture or scale. All four evaluated MLLMs achieve higher exam scores with the system, demonstrating its broad and reliable effectiveness. As illustrated in Fig.~\ref{fig:score_ipho_2025}, \textsc{PhysicsMinions} consistently boosts Intern-S1's scores across all problems on the latest IPhO, with the most notable improvement in the Text+Data Figure modality, underscoring its strong ability to enhance multimodal reasoning.

\textbf{(2) \textsc{PhysicsMinions} enables open-source MLLMs to achieve substantial medal progression—from bronze to silver, and from silver to gold.}  
For instance, Qwen2.5VL-32B-Instruct, a relatively weaker reasoner, improved from zero silvers to five silvers. Intern-S1 advanced from 2 golds, 4 silvers, and 1 bronze in the single-model setting to 6 golds and 1 silver with \textsc{PhysicsMinions}. Likewise, InternVL3.5-241B-A28B with \textsc{PhysicsMinions} also reached 6 golds and 1 silver, with remarkable leaps from bronze to gold in IPhO and PanPhO. Notably, this work marks \textbf{the first time} that open-source MLLMs have achieved a gold medal in the latest IPhO, including both Intern-S1 and InternVL3.5-241B-A28B, underscoring how the coevolutionary system, through verification and reflection, can elevate open-source reasoning to the Olympiad gold level.

\textbf{(3) \textsc{PhysicsMinions} pushes the boundary of the closed-source MLLM, surpassing the top single model and even human contestants in several Olympiads.}  
With \textsc{PhysicsMinions}, Gemini-2.5-Flash-Thinking became the first model to win gold medals in all seven Olympiads. Notably, while no single model on the HiPhO leaderboard had ever reached gold in the latest EuPhO, the system accomplished this milestone. It also outperformed the best single-model scores on three exams—APhO, EuPhO, and F=MA. In mechanics, the gains were especially striking: the system scored 24/25 on F=MA (25 multiple-choice questions), nearly perfect, and even surpassed human contestants in PanMechanics, demonstrating unprecedented advances in mechanical reasoning.

In summary, \textsc{PhysicsMinions} consistently boosts performance, enables medal breakthroughs for open-source MLLMs, and pushes closed-source models beyond prior limits, showcasing its potential for advancing multimodal physical reasoning at Olympiad scale.

\subsection{Comparison with Other Frameworks}
\label{sec:other_framework}

We compare \textsc{PhysicsMinions} with three representative baselines: Best-of-N, Self-Mixture-of-Agents (Self-MoA), and Self-Refine. Best-of-N \citep{stiennon2020learning} selects the output with the highest total exam score from $N$ independent runs. Self-MoA \citep{li2025rethinking} ensembles diverse outputs from a single model by varying prompting strategies and then aggregates them into a final solution. Self-Refine \citep{madaan2023self} iteratively prompts the model to critique and improve its own answers. Further details and discussion of these methods are provided in Appendix~\ref{appendix:framework}.

\begin{table}[t]
\centering
\caption{Performance comparison of different frameworks on 7 latest physics Olympiads using the exam score metric. \medalbox{Gold!50}{Gold}, \medalbox{Silver!70}{Silver}, and \medalbox{Bronze!40}{Bronze} indicate scores above the respective thresholds. \textbf{Bold} marks the highest score, with \textsc{PhysicsMinions} achieving the best results in all settings.}
\label{tab:other_framework}
\small
\resizebox{\textwidth}{!}{%
\setlength{\tabcolsep}{2.5pt}
\begin{tabular}{lccccccc|ccc}
\toprule
\textbf{Latest Physics Olympiads} & \textbf{IPhO} & \textbf{APhO} & \textbf{EuPhO} & \textbf{NBPhO} & \textbf{PanPhO} & \textbf{PanMechanics} & \textbf{F=MA} & \multicolumn{3}{c}{\textbf{Medal}} \\ 
\midrule
\rowcolor{Gold!40}
Gold Medal   & 19.7 & 23.3 & 16.5 & 28.6 & 41.5 & 52.0 & 15.0 & \multicolumn{3}{>{\cellcolor{white}}l}{\textcolor{Gold}{\faMedal}} \\
\rowcolor{Silver!60}
Silver Medal & 12.1 & 18.7 & 9.8  & 20.1 & 28.5 & 36.0 & 11.0 & \multicolumn{3}{>{\cellcolor{white}}c}{\textcolor{Silver}{\faMedal}} \\
\rowcolor{Bronze!40}
Bronze Medal & 7.2  & 13.1 & 5.8  & 15.2 & 14.5 & 20.0 & 9.0  & \multicolumn{3}{>{\cellcolor{white}}r}{\textcolor{Bronze}{\faMedal}} \\
\midrule

Intern-S1 & \cellcolor{Silver!60}15.9 & \cellcolor{Silver!60}21.7 & \cellcolor{Bronze!40}9.0 & \cellcolor{Silver!60}23.0 & \cellcolor{Silver!60}41.1 & \cellcolor{Gold!35}60.4 & \cellcolor{Gold!35}18.4 & \mc{Gold}{2} & \mc{Silver}{4} & \mc{Bronze}{1} \\

\;\;\textbf{+ \textsc{PhysicsMinions}} & \cellcolor{Gold!35}\textbf{20.8} & \cellcolor{Gold!35}\textbf{25.2} & \cellcolor{Silver!60}\textbf{10.1} & \cellcolor{Gold!35}\textbf{28.9} & \cellcolor{Gold!35}\textbf{46.8} & \cellcolor{Gold!35}\textbf{68.7} & \cellcolor{Gold!35}\textbf{22.7} & \mc{Gold}{6} & \mc{Silver}{1} & \mc{Bronze}{0} \\

\;\;+ Best-of-N ($N=3$) & \cellcolor{Silver!60}16.6 & \cellcolor{Silver!60}22.9 & \cellcolor{Bronze!40}9.7 & \cellcolor{Silver!60}25.2 & \cellcolor{Gold!35}46.0 & \cellcolor{Gold!35}68.5 & \cellcolor{Gold!35}21.0 & \mc{Gold}{3} & \mc{Silver}{3} & \mc{Bronze}{1} \\

\;\;+ Self-MOA & \cellcolor{Silver!60}15.8 & \cellcolor{Silver!60}21.7 & \cellcolor{Silver!60}9.9 & \cellcolor{Bronze!40}16.9 & \cellcolor{Gold!35}42.7 & \cellcolor{Gold!35}68.5 & \cellcolor{Gold!35}19.0 & \mc{Gold}{3} & \mc{Silver}{3} & \mc{Bronze}{1} \\

\;\;+ Self-Refine & \cellcolor{Silver!60}18.6 & \cellcolor{Gold!35}24.6 & \cellcolor{Bronze!40}9.0 & \cellcolor{Silver!60}28.4 & \cellcolor{Silver!60}37.8 & \cellcolor{Gold!35}59.0 & \cellcolor{Gold!35}21.0 & \mc{Gold}{3} & \mc{Silver}{3} & \mc{Bronze}{1} \\
\bottomrule
\end{tabular}%
}
\end{table}

As shown in Table~\ref{tab:other_framework}, \textsc{PhysicsMinions} exhibits clear advantages in both consistent score improvement and medal progression. Its coevolutionary system enhances performance more reliably than alternative frameworks. In contrast, Self-MoA can produce incorrect solutions when ensembling candidates, as illustrated by Intern-S1 scoring lower than the single model in NBPhO. Self-Refine, while employing verification and reflection, lacks the dual-stage verification and coevolutionary reflection of \textsc{PhysicsMinions}, resulting in lower scores in PanPhO and PanMechanics. In terms of medal outcomes, \textsc{PhysicsMinions} elevates Intern-S1 from two to six golds, including at IPhO, whereas other frameworks achieve only three. Furthermore, \textsc{PhysicsMinions} surpasses the Best-of-N performance, indicating it breaks through the performance ceiling of individual models.

\subsection{Scaling Performance under Pass@$k$}

\begin{figure}[H]
    \centering
    \includegraphics[width=0.48\linewidth]{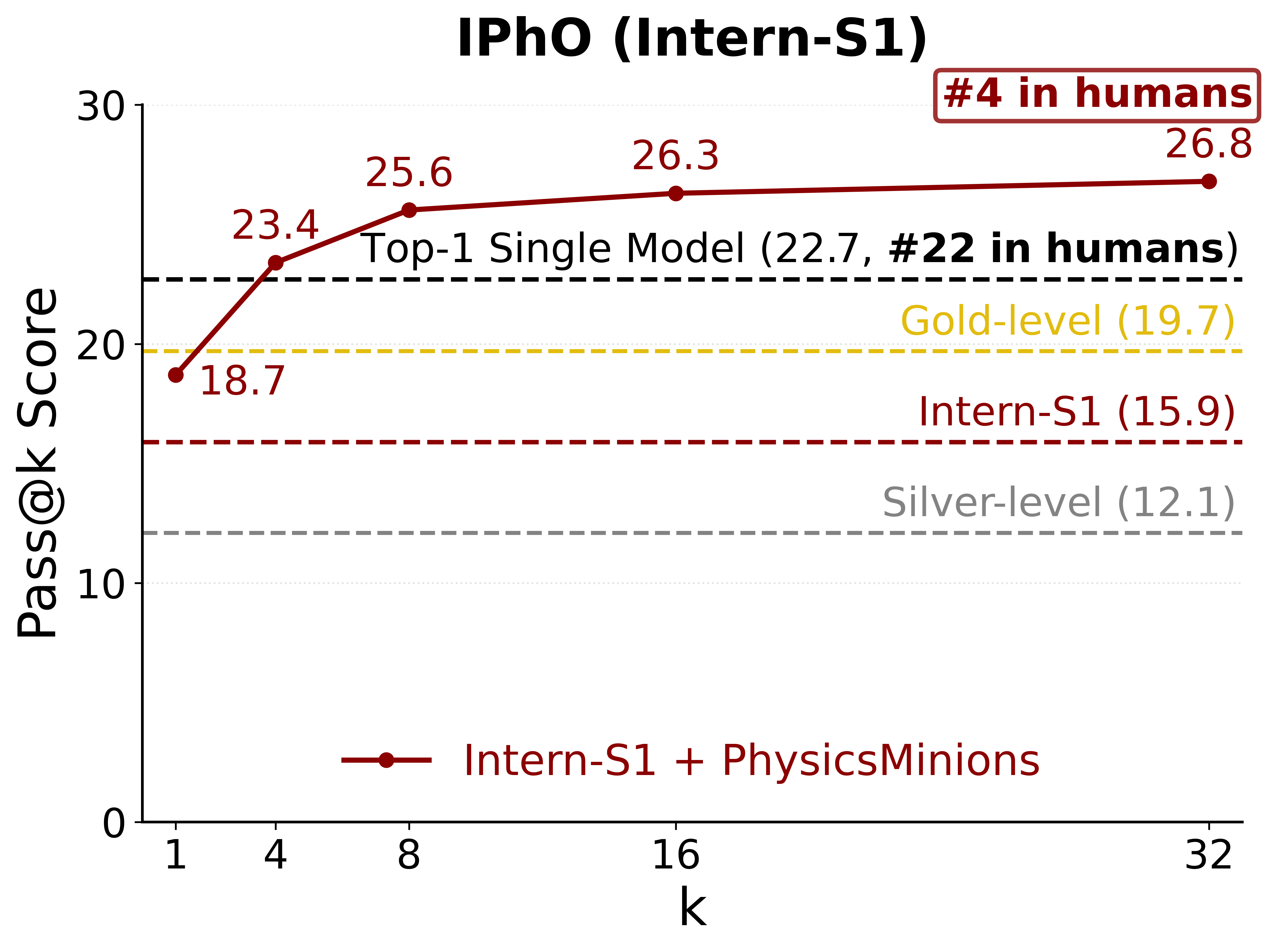}
    \includegraphics[width=0.48\linewidth]{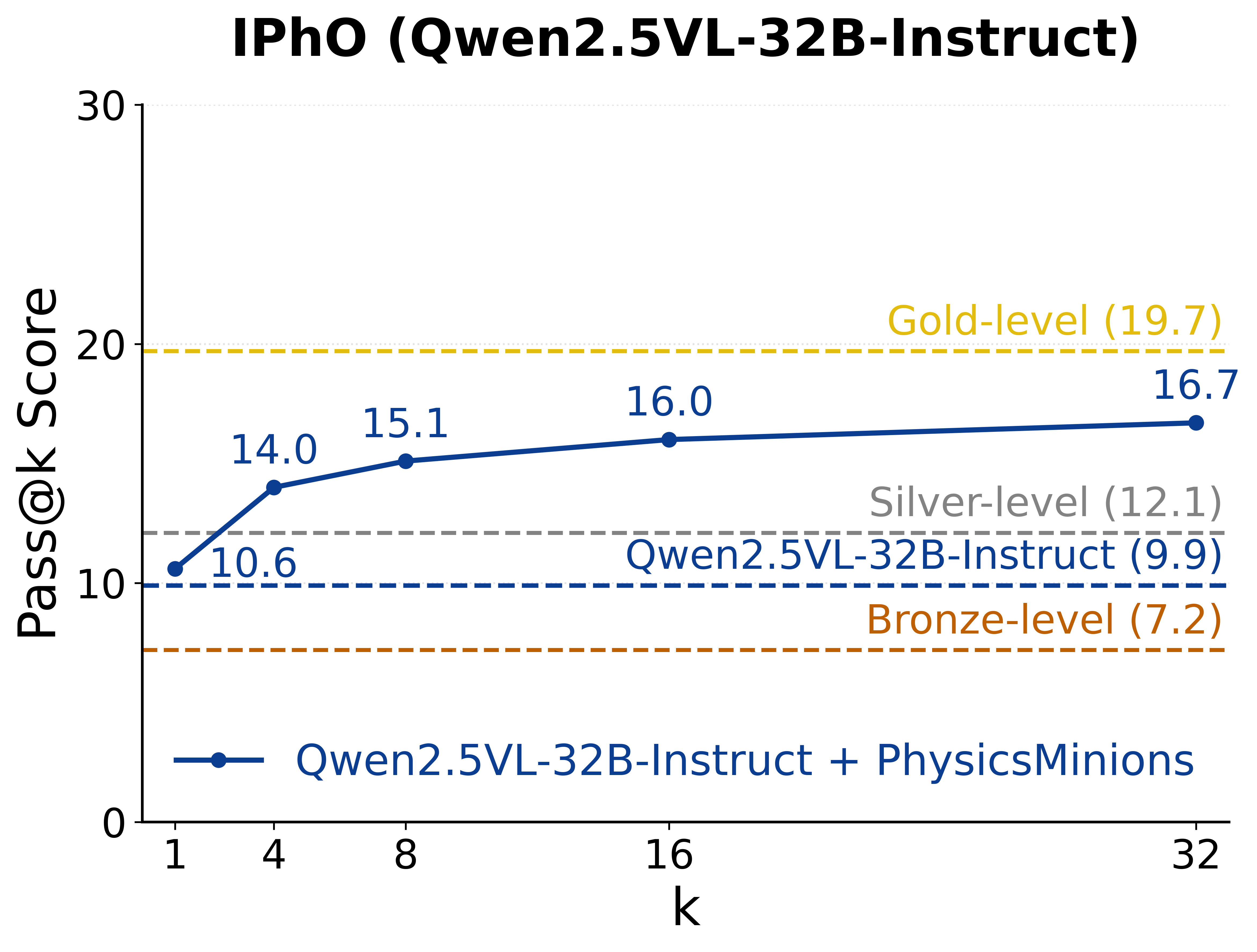}
    \vspace{-1mm}
    
    \caption{Scaling performance of Intern-S1 and Qwen2.5VL-32B-Instruct on the latest IPhO.}
    \label{fig:ipho2025_passk}
\end{figure}

Pass@$k$ evaluates a model's best score over $k$ independent attempts by taking the highest-scoring solution per problem. Fig.~\ref{fig:ipho2025_passk} illustrates the scaling behavior of \textsc{PhysicsMinions}, revealing three key observations. \textbf{(1) Continuous performance evolution:} Intern-S1+\textsc{PhysicsMinions} improves steadily from Pass@1 to Pass@4 with a 4.7-point gain, surpassing the top single-model score of 22.7 (22$^\text{nd}$ among humans). At Pass@32, it reaches 26.8/30, \textbf{ranking 4$^\textbf{th}$ of 406 contestants} and surpassing 99\% of human contestants, with further growth potential. \textbf{(2) Early breakthroughs:} significant progress is achieved at Pass@4, where Intern-S1 upgrades from silver to gold, and Qwen2.5VL-32B-Instruct rises from bronze to silver. \textbf{(3) Base model determines ceiling:} the intrinsic capability of the base model constrains the system's upper bound. Stronger models, such as Intern-S1, benefit more from the coevolutionary system, yielding larger performance gains.
\vspace{-1mm}

\subsection{Ablation Studies and Hyperparameter Analysis}
\label{sec:ablation}

\begin{figure}
    \centering
    \includegraphics[width=1\linewidth]{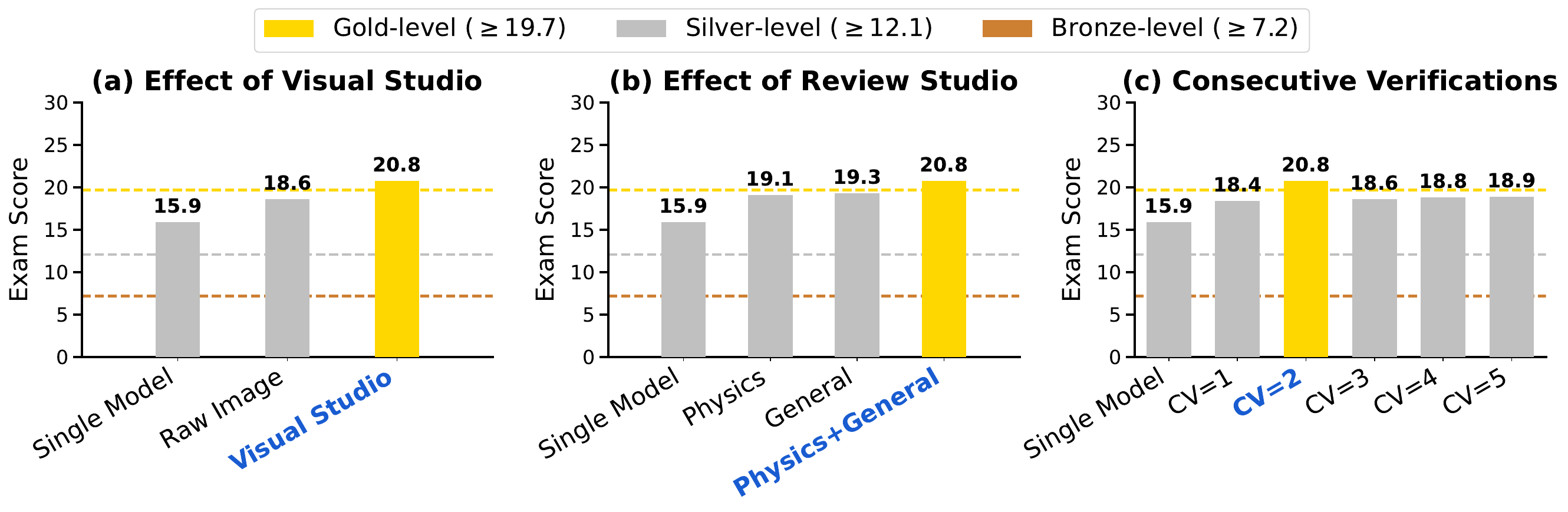}
    \caption{Ablation study and hyperparameter analysis using Intern-S1 on the latest IPhO.}
    \label{fig:ablation}
\end{figure}

\textbf{Effect of Visual Studio.}  
Visual Studio converts diagrams and plots into structured JSON descriptions (see Fig.~\ref{fig:framework}), which we compare against directly using raw images. As shown in Fig.~\ref{fig:ablation}(a), Visual Studio achieves higher scores, since structured representations provide explicit cues that support reasoning. In contrast, raw images are harder for the solver to interpret directly, making it difficult to extract quantitative details and leading to weaker performance.

\textbf{Effect of Review Studio.} Review Studio employs a dual-stage verification with a Physics-Verifier followed by a General-Verifier. We compare each verifier used individually against their combination. As shown in Fig.~\ref{fig:ablation}(b), the combination yields the highest score of 20.8, whereas the Physics-Verifier alone only checks physics consistency and the General-Verifier alone lacks domain-specific rigor. Together, the two verifiers complement each other to achieve gold-level performance.

\textbf{Effect of Consecutive Verifications.}  
The key hyperparameter in the system is the number of consecutive verifications (CV), where a candidate solution is accepted only if it passes CV checks in a row. On the latest IPhO, we tested $\text{CV} \in \{1,2,3,4,5\}$ and observed the score rises from 15.9 to 20.8 at $\text{CV}=2$, while larger values may trigger overthinking and reduce scores. As CV increases, token consumption increases accordingly, reaching about 1.9× and 3.2× that of $\text{CV}=2$ for $\text{CV}=3$ and $\text{CV}=5$, respectively (measured on IPhO Q3-A6 as an example). Thus, we adopt $\text{CV}=2$ as an empirically efficient setting, though the optimal value may vary with problem difficulty or model, with its effectiveness confirmed in our experiments.
\section{Discussion}

\subsection{Case Study on Dual-stage Verification}

\begin{figure}[htbp]
    \centering
    \includegraphics[width=0.92\linewidth]{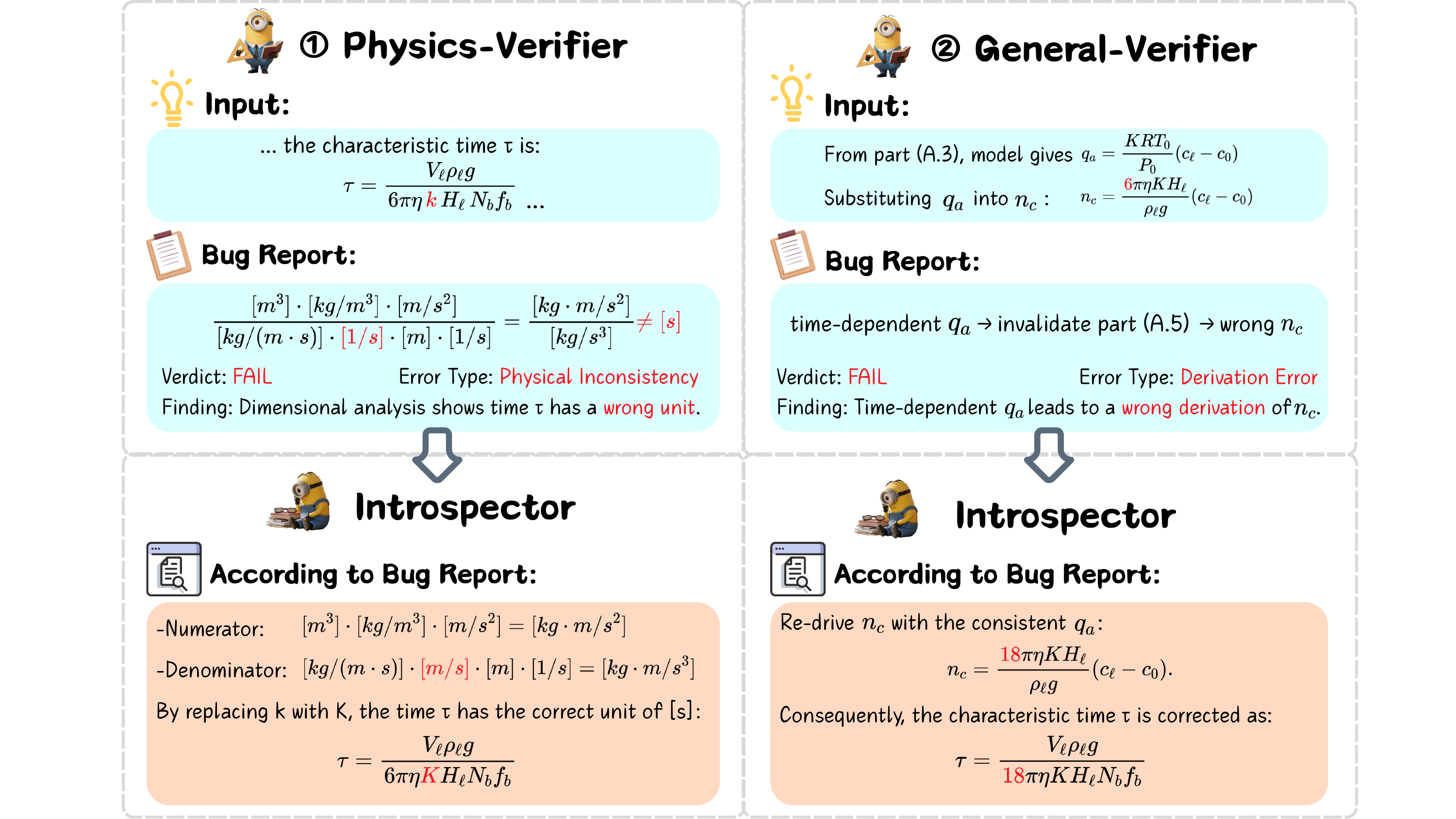}
    \caption{Case study of dual-stage verification on IPhO Q3-A6 using Intern-S1. The single model achieves only 0.2 points, whereas \textsc{PhysicsMinions}, with dual-stage verification and self-reflection, attains the full score of 1.1 points, demonstrating substantial improvement.}
    \label{fig:casestudy}
\end{figure}

As illustrated in Fig.~\ref{fig:casestudy}, we present a case study showing how dual-stage verification refines solutions. \textbf{(1) The first stage:} The Physics-Verifier conducts domain-specific checks on units, conversions, and physical consistency, identifying that the characteristic time $\tau$ has an incorrect unit under dimensional analysis. Guided by this bug report, the introspector corrects the error by substituting the proper physical variable. \textbf{(2) The second stage:} The General-Verifier evaluates logical consistency, derivation soundness, and numerical validity. It detects a derivation error that a time-dependent $q_a$ leads to a flawed derivation of $n_c$. The introspector then re-derives $n_c$ and arrives at the correct solution with full credit. This case underscores the essential role of dual-stage verification in our coevolutionary system: the capacity to detect errors is the foundation of iterative self-improvement, and without reliable error detection, repeated reflection cannot converge on the correct answer.


\subsection{Limitations}
\label{sec:limitation}

\begin{figure}[htbp]
    \centering
    \includegraphics[width=0.8\linewidth]{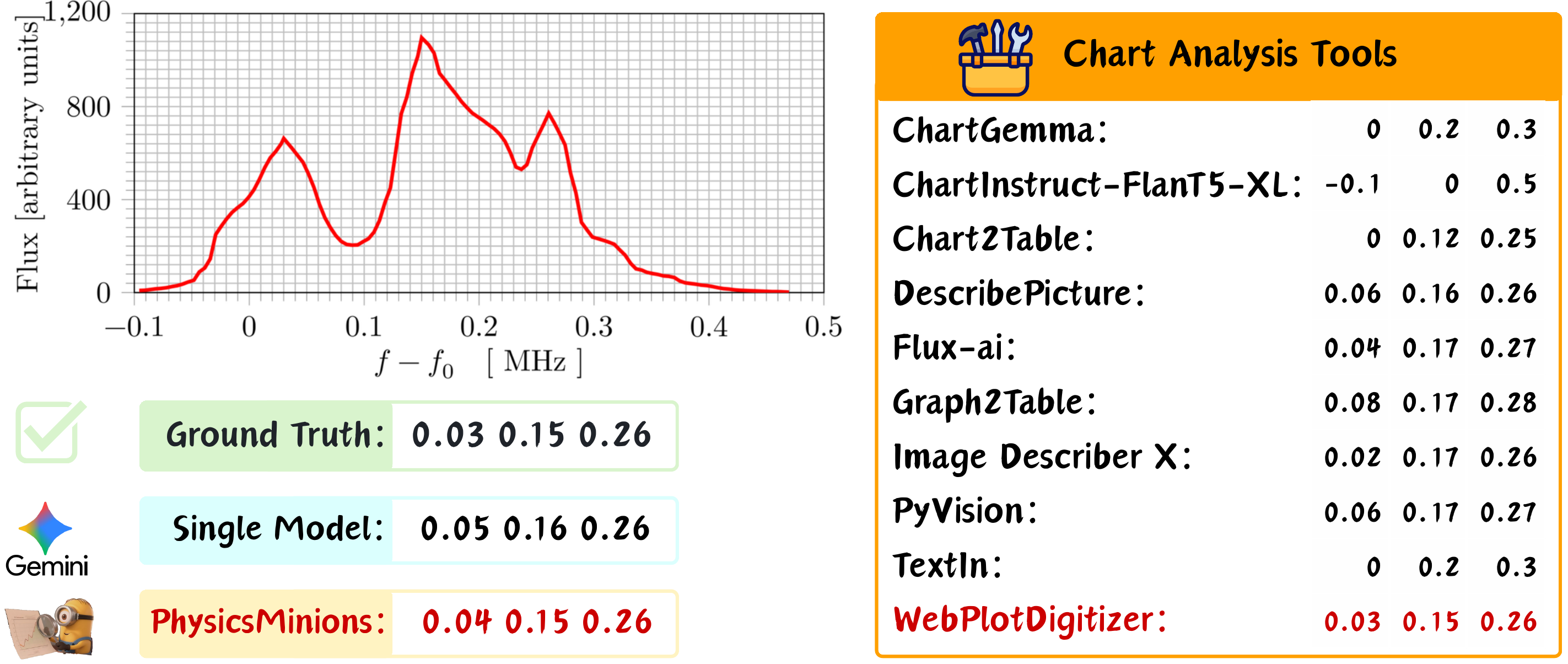}
    \caption{Limitation of \textsc{PhysicsMinions}' Visual Studio in precise data extraction. Example: IPhO Q1-C4 requires identifying the x-coordinates of all three peaks in the curve.}
    \label{fig:limitation}
\end{figure}

Visual Studio significantly improves multimodal reasoning, yet precise data extraction remains challenging. For instance, in Fig.~\ref{fig:limitation}, a chart with three peaks is partially misinterpreted: \textsc{PhysicsMinions} correctly extracts two values, outperforming a single model, but still selects one peak incorrectly. Besides, we evaluated various chart analysis tools \citep{masry2024chartgemma,masry2024chartinstruct,pp-chart2table,describepicture,fluxai_describe_image,graph2table,image_describer_x,zhao2025pyvision,textin_pdf_to_markdown}, most of which exhibit worse recognition capabilities than the single Gemini model, while only WebPlotDigitizer \citep{automeris_webplotdigitizer_home} achieves fully correct extraction but requires human-in-the-loop operations, making it impractical for automated systems. These results underscore the need to develop fully automated, high-precision visual extraction capabilities to further enhance multimodal reasoning performance. Illustrations of tools are provided in Appendix~\ref{appendix:tool}.

\section{Conclusion and Future Work}

Physics underpins our ability to shape the real world, and physics Olympiads distill this challenge into rigorous tests that expose the limits of single-model approaches. To address this, we proposed \textsc{PhysicsMinions}, a coevolutionary multimodal multi-agent system designed to push beyond the single-model ceiling. By integrating a Visual Studio for structured perception, a Logic Studio for iterative solution refinement, and a Review Studio for dual-stage verification, the framework evolves solutions through continuous critique and feedback. Evaluated on 7 latest physics Olympiads, it delivers historic breakthroughs, including the first open-source gold in the latest IPhO and a Pass@32 score of 26.8/30 that ranks 4$^\text{th}$ among 406 contestants, surpassing 99\% of human contestants.

Future work will focus on three directions: (i) enhancing visual understanding and multimodal perception in the Visual Studio, (ii) expanding tool use with external solvers and domain-specific verifiers to strengthen correction, and (iii) extending the coevolutionary paradigm to other Olympiad-level domains beyond physics.

\subsubsection*{Acknowledgments}

This work was supported by the Shanghai Artificial Intelligence Laboratory and a locally commissioned task from the Shanghai Municipal Government.

\bibliography{ref}
\bibliographystyle{iclr2025_conference}

\newpage
\appendix

\lstset{
  basicstyle=\ttfamily\footnotesize,
  columns=flexible,
  breaklines=true,
  breakatwhitespace=true,
  frame=none,                     
  rulecolor=\color{gray!30},
  backgroundcolor=,               
  numbers=none,
  xleftmargin=1em,
  xrightmargin=1em,
  aboveskip=0.5em,
  belowskip=0.5em,
  literate={—}{--}1
           {↔}{{$\leftrightarrow$}}1
           {°}{\textdegree}1
           {β}{{$\beta$}}1
           {ħ}{{$\hbar$}}1
           {≈}{{$\approx$}}1
           {−}{-}1
           {–}{-}1
           {→}{->}1
}

\begin{center}
    \textbf{\Large Supplemental Material of \textsc{PhysicsMinions}}
\end{center}

This document provides supplementary material to complement the main paper. It includes detailed descriptions of the system, prompts, comparative frameworks, additional results, and analysis tools. Specifically:

\begin{itemize}
    \item \textbf{Appendix~\ref{appendix:llm}} describes how large language models were used in this work.
    
    \item \textbf{Appendix~\ref{appendix:system}} presents a detailed description of the \textsc{PhysicsMinions} system, including:
    \begin{itemize}
        \item Appendix~\ref{appendix:system_image}: Image processing pipeline
        
        \item Appendix~\ref{appendix:system_coevolutionary}: Coevolutionary iteration strategy
        
        \item Appendix~\ref{appendix:system_implementation}: Implementation details
    \end{itemize}

    \item \textbf{Appendix~\ref{appendix:prompt}} provides the complete set of prompts used in \textsc{PhysicsMinions}, including:
    \begin{itemize}
        \item Appendix~\ref{appendix:prompt_visual}: Prompts of Visual Studio
        
        \item Appendix~\ref{appendix:prompt_logic}: Prompts of Logic Studio
        
        \item Appendix~\ref{appendix:prompt_review}: Prompts of Review Studio
    \end{itemize}

    \item \textbf{Appendix~\ref{appendix:framework}} introduces the comparative frameworks considered in our experiments, including:
    \begin{itemize}
        \item Appendix~\ref{appendix:framework_bestofn}: Best-of-N strategy
        
        \item Appendix~\ref{appendix:framework_selfmoa}: Self-MoA
        
        \item Appendix~\ref{appendix:framework_selfrefine}: Self-Refine

        \item Appendix~\ref{appendix:framework_advantages}: Advantages over comparison frameworks
    \end{itemize}

    \item \textbf{Appendix~\ref{appendix:results}} provides the description of physics Olympiads and additional results, including:
    \begin{itemize}
        \item Appendix~\ref{appendix:results_overview}: Overview of physics Olympiads
        
        \item Appendix~\ref{appendix:results_modality}: Performance gains across modality types
        
        \item Appendix~\ref{appendix:results_field}: Performance gains across physics fields
    \end{itemize}

    \item \textbf{Appendix~\ref{appendix:tool}} gives an overview of the chart analysis tools tested in Section~\ref{sec:limitation}.
\end{itemize}

\section{The Use of Large Language Models}
\label{appendix:llm}

The paper was originally drafted by the authors, with GPT-5 used only for polishing language and grammar. In Fig.~\ref{fig:overview}, the Minion-style concept and overall layout were designed by the authors; four standalone images were generated with Google Nano Banana\footnote{\url{https://www.nano-banana.ai}} under author-provided instructions. In Fig.~\ref{fig:framework}, only the Minion-style icons were generated with Doubao\footnote{\url{https://www.doubao.com/chat/}} based on author-provided instructions, while the framework itself was created by the authors. The evaluation pipeline follows the HiPhO benchmark \citep{2025hipho}, using Gemini-2.5-Flash as the grader for model-based evaluation. No other substantive use of LLMs was involved in the ideation of this paper.

\section{A Coevolutionary System of \textsc{PhysicsMinions}}
\label{appendix:system}

\subsection{Image Processing Pipeline}
\label{appendix:system_image}

The system incorporates a dedicated image processing pipeline for handling image inputs:

\begin{enumerate}
    \item \textbf{Image Reading:} Extracts initial visual information from the input image.
    
    \item \textbf{Image Improvement:} Improves the interpretation of the extracted image content.
    
    \item \textbf{Image Verification:} Validates the interpretation by re-checking against the original image.
    
    \item \textbf{Multi-Round Refinement:} Iteratively refines the interpretation until it consistently passes verification.
    
    \item \textbf{Consecutive Verification:} Accepts an interpretation only after it passes the required number of consecutive verifications.
\end{enumerate}


\subsection{Coevolutionary Iteration Strategy}
\label{appendix:system_coevolutionary}

The \textsc{PhysicsMinions} employs a multi-round coevolutionary iteration strategy:

\begin{enumerate}
    \item \textbf{Initial Solution:} The Solver generates an initial solution, which is improved by the Introspector to correct errors.
    
    \item \textbf{Dual-Stage Verification:}
    \begin{itemize}
        \item \textbf{Stage 1:} The Physics-Verifier checks domain-specific physics consistency (e.g., units, constants, assumptions).  
        \item \textbf{Stage 2:} The General-Verifier checks detailed logical and computational correctness.
    \end{itemize}
    
    \item \textbf{Iterative Refinement:} 
    \begin{itemize}
        \item \textbf{Failure:} If either Physics-Verifier or General-Verifier fails, a bug report is generated, corrected by the Introspector, and re-verified.
        \item \textbf{Success:} If both pass, the count of consecutive successes increases; once the threshold is met, the solution is accepted, otherwise verification repeats.
    \end{itemize}
    
    \item \textbf{Consecutive Verification:} A solution is accepted only after meeting the predefined threshold of consecutive successful verifications. If the threshold of consecutive failures is reached, the process restarts with a new initial solution.
\end{enumerate}


\subsection{Implementation Details}
\label{appendix:system_implementation}

\begin{algorithm}[H]
\caption{PhysicsMinions: A Coevolutionary Multimodal Multi-Agent System}
\begin{algorithmic}[1]
    \Require Problem instance (incl.\ images); Consecutive Verification threshold $\mathrm{CV}$ (default: $2$)
    \Ensure Final solution
    \State $I \gets \text{VisualExtract}(\text{Problem})$ \Comment{Visual Studio: extract structured visual info}
    \State $S \gets \text{GenerateInitialSolution}(\text{Problem}, I)$ \Comment{Solver uses structured input $I$}
    \State $c \gets 0$; $f \gets 0$ \Comment{$c$=consecutive successes, $f$=consecutive failures}
    \While{not converged} \Comment{Loop at most 5 iterations}
        \State $S \gets \text{IntrospectorImprove}(S)$
        \State $\text{Pass or Fail, Bug\_report} \gets \text{PhysicsVerify}(S)$ \Comment{Stage 1: Physics-Verifier}
        \If{Fail}
            \State $S \gets \text{IntrospectorImprove}(\text{Bug\_report})$
            \State $c \gets 0$; $f \gets f+1$ \Comment{reset success; increment failures}
            \If{$f \ge \mathrm{CV}$}
                \State $S \gets \text{GenerateInitialSolution}(\text{Problem},I)$
                \State $c \gets 0$; $f \gets 0$
            \EndIf
            \State \textbf{continue}
        \EndIf
        \State $\text{Pass or Fail, Bug\_report} \gets \text{GeneralVerify}(S)$ \Comment{Stage 2: General-Verifier}
        \If{Fail}
            \State $S \gets \text{IntrospectorImprove}(\text{Bug\_report})$
            \State $c \gets 0$; $f \gets f+1$
            \If{$f \ge \mathrm{CV}$}
                \State $S \gets \text{GenerateInitialSolution}(\text{Problem},I)$
                \State $c \gets 0$; $f \gets 0$
            \EndIf
            \State \textbf{continue}
        \Else
            \State $c \gets c+1$; $f \gets 0$
            \If{$c \ge \mathrm{CV}$}
                \State \Return $S$
            \EndIf
        \EndIf
    \EndWhile
\end{algorithmic}
\end{algorithm}

\newpage
\section{Detailed Prompts of \textsc{PhysicsMinions}}
\label{appendix:prompt}

The complete prompts are included in the supplementary material.

\subsection{Prompts of Visual Studio}
\label{appendix:prompt_visual}

\begin{itemize}
    \item \textbf{Inspector:} extracts initial structured information from the image.  
    
    \item \textbf{Introspector (Image):} improves and refines the structured information.  

    \item \textbf{Verifier (Image):} validates the extracted information against the original image.
\end{itemize}

\begin{tcolorbox}[
  breakable,
  title={Inspector},
  fonttitle=\bfseries,
  colback=pale!10,           
  colbacktitle=pale,         
  boxrule=0.5pt,
  arc=2mm
]
\begin{lstlisting}
# Your Task
Analyze physics problem's **reference figures** (schematic, plot, free-body diagram, circuit diagram, optical diagram, waveform, table, object image, or combination). Extract **every possible visual detail** without fabrication, producing a **structured JSON summary** that would allow accurate reconstruction of the figure.

## Step 1 – Coarse Scan (Global Understanding)
For each figure:
1. Identify the **overall purpose** and **theme** of the figure.
2. Locate and read any **title, caption, or source**.
3. Count and identify **subfigures** (a), (b), (c)...
- For each, note:
    - **Theme** (main concept)
    - **Figure type** (choose from: `plot`, `free_body`, `circuit`, `optics`, `waveform`, `table`, `schematic`, `object_image`, `other`)

### Classification Priority Rules
- If a figure clearly matches **circuit diagram**, **optical setup/diagram**, or **free-body diagram**,  
→ **must** classify it as `circuit`, `optics`, or `free_body` respectively,  
→ **do NOT** lump these into `schematic`.
- Use `schematic` **only for general physical setups / abstract line drawings** that do not fit into the above categories.
- Use `object_image` if it is a **real-world photo or realistic rendering** of experimental apparatus or objects.
- If the figure does not fit any known category, classify as `other` and provide a short explanation.

## Step 2 – Detailed Scan (Type-Specific Rules)
For **each** figure/subfigure, follow **type-specific extraction rules**:

### A. Axis-Based Figures (`figure_type = "plot"`)
Extract enough information to rebuild the plot.

- **x_axis / y_axis**:
- `label`: From axis text (or `"unknown"` if missing)
- `unit`: Prefer SI units (or `"unknown"`)
- `range`: `[min, max]` estimated from labeled tick marks
- `scale`: `"linear"` or `"log"`
- `ticks`: List **all labeled major tick values** as numbers or strings exactly as shown. If a label is not numeric, keep its string.

#### Curves  — REQUIRED `curves[*].data` FORMAT UPDATED
For each curve, extract using the schema below. **Do not fabricate** values; if a y-value cannot be read, set it to `"unknown"` but still include the x tick.

- `name`: Legend label or `"unknown"`
- `color`: Visible curve color (e.g., `"blue"`, `"red"`, or hex if distinctive)
- `line_style`: `"solid"`, `"dashed"`, `"dotted"`, etc.
- `overall_trend`: A 1–3 sentence summary of the curve's behavior over the full x-range (e.g., increasing then plateau; single peak near x≈2; S-shaped with inflection).
- `data`:
- `by_x_ticks`: **Must include**:
    1. **All visible labeled major x ticks**.
    2. **Starting point** (leftmost visible data).
    3. **Ending point** (rightmost visible data).  
    Each entry format:
    ```json
    { "x": <value>, "y": <estimated_or_"unknown">, "read_method": "from_graph", "note": "<optional short note>" }
    ```
- `special_points`: **Must include all identifiable notable points**, e.g.:
    ```json
    { "type": "peak" | "valley" | "inflection" | "x_intercept" | "y_intercept" | "endpoint_feature" | "intersection",
    "x": <value_or_"unknown">, "y": <value_or_"unknown">, "with_curve": "<name if intersection>",
    "description": "<why it's special>", "confidence": "high"|"medium"|"low" }
    ```

#### Scatter Points
For each scatter series, extract using the schema below. **Do not fabricate** values; if a y-value cannot be read, set it to `"unknown"`.

- `name`: Series label or `"unknown"`
- `color`: Marker color
- `shape`: Marker shape (circle, square, etc.)
- `overall_distribution`: 1–3 sentence description of how the points are arranged (clustered, linear trend, random spread, etc.)
- `data`:
- `by_x_ticks`: **Must include**:
    1. **All visible labeled major x ticks** (0, 1, 2...).
    2. **Uniformly sampled intermediate points** (pick representative points in between ticks if present).
    3. **Starting point cluster** (leftmost x with points).
    4. **Ending point cluster** (rightmost x with points).  
    Each entry format:
    ```json
    { "x": <tick_value>, "y_values": [<all points near this tick or "unknown">], "read_method": "from_graph", "note": "<optional>" }
    ```
- `special_points`: **Must include all identifiable notable points**, such as outliers, cluster centers, gaps, or notable trend points:
    ```json
    { "type": "outlier" | "cluster_center" | "gap" | "trend_point",
    "x": <value_or_"unknown">, "y": <value_or_"unknown">,
    "description": "<why it's special>", "confidence": "high"|"medium"|"low" }
    ```

### B. Free-Body Diagrams (`free_body`)
- List the object(s)
- All forces acting (gravity, normal, friction, tension, applied forces)
- Force directions and decompositions
- Equilibrium hints

### C. Circuit Diagrams (`circuit`)
- All components with values and symbols
- Polarities and connections
- Switch states, measurement points

### D. Optical Diagrams (`optics`)
- Media (air, glass, water, etc.)
- Rays (incident, reflected, refracted)
- Lens/mirror types and focal lengths
- Object/image positions

### E. Waveforms (`waveform`)
- Amplitude, period, frequency, phase
- Envelope shapes, modulation patterns
- Time base scale

### F. Tables (`table`)
- Column names and units
- All values
- Missing-data markers

### G. Schematics / Physical Setups (`schematic`)
1) **Elements to Extract (must be complete but concise)**
- **Points**: list all labeled points (every symbol shown).
- **Vectors**: include all vectors with start->end notation (e.g., `S->Sv`), preserve arrow style/color if meaningful.
- **Segments**: use `start-end` format (e.g., `S-E`, `s-S`). If segment length/value is known, include it as `"value"`.
- **Angles**: include only angles with labels (e.g., β) or known values (e.g., right angles). Label them as `angleABC` (e.g., `angleCET`).
- **Arcs**: include arcs with center and through point (e.g., `"center": "C", "through": "E"`), specify dashed/solid if relevant.
- **Styles**: only if meaningful (color, dashed, arrow).

2) **Output Format (JSON-like)**
- **Labeling rule**:
- Use the exact label from the diagram if shown.
- If none, use `start-end` for segments and `start->end` for vectors.
- For angles, always use `angleABC` form, where B is the vertex.

### H. Object Images (`object_image`)
- Objects present
- Shape, composition, materials
- Spatial relationships and distances

---
# Step 3  Output Rules
- **If single figure**:
```json
{
"title": "",
"figure_type": "",
"x_axis": {...},
"y_axis": {...},
"curves": [...],
"scatter_points": [...]
}
```
- **If multiple figures**:
{
"title": "",
"figure": [
    {
    "id": "a",
    "theme": "",
    "figure_type": "",
    "x_axis": {...},
    "y_axis": {...},
    "curves": [...],
    "scatter_points": [...]
    },
    {
    "id": "b",
    ...
    }
]
}
\end{lstlisting}
\end{tcolorbox}

\begin{tcolorbox}[
  breakable,
  title={Introspector (Image)},
  fonttitle=\bfseries,
  colback=pale!10,           
  colbacktitle=pale,         
  boxrule=0.5pt,
  arc=2mm
]
\begin{lstlisting}
You are an IMAGE-READING EXTRACTOR for physics-related figures (plots, circuit diagrams, schematics, object photos).

Goal:  
Read the IMAGE ONLY and output a FULL JSON in the EXACT SAME SCHEMA required by read_image_prompt.  
Your output must be self-contained, consistent, and reflect the visible content of the image.

Critical rules:
- Follow EXACTLY the JSON schema used by read_image_prompt (keys, nesting, arrays).  
- Use the IMAGE ONLY. Do not fabricate elements that are not visible.  
- If something is unclear/occluded, include it with lower confidence and describe uncertainty briefly.  
- Keep the same JSON structure:
{
    "title": <str>,
    "figure": [
    { ... sub-figure #1 ... },
    { ... sub-figure #2 ... }
    ]
}
- For each sub-figure, set an appropriate figure_type: "plot" | "schematic" | "circuit" | "photo" (or other valid types defined in read_image_prompt).

Main detection & extraction focus:
1. **Theme**: Ensure the high-level description matches the figure (e.g., is it a plot, circuit, geometry, or photo).  
2. **Elements**: Identify and record points, vectors, segments, angles, arcs, circuit components, arrows, right-angle marks, hats/circles/subscripts/superscripts, geometric constraints (perpendicular/parallel).  
3. **Data**:  
- For plots: re-read axis labels, units, ranges, scales, ticks.  
- Capture curves with name, color, line_style, overall_trend, compact data samples (e.g., by_x_ticks).  
- Identify special_points (peaks, valleys, intercepts) with accurate values and confidence.  
- Ensure numbers and symbols are correct and consistent.  
4. **Labels**: Check that Greek letters, subscripts, superscripts, and vector/matrix notation match the figure text exactly.

Consistency & quality:
- Cross-check that labels and values remain consistent across all sections.  
- Use numeric precision appropriate to image resolution; if approximate, indicate so with "confidence".  
- If a figure type or field is not present in the image, omit it rather than hallucinating.
\end{lstlisting}
\end{tcolorbox}

\begin{tcolorbox}[
  breakable,
  title={Verifier (Image)},
  fonttitle=\bfseries,
  colback=pale!10,           
  colbacktitle=pale,         
  boxrule=0.5pt,
  arc=2mm
]
\begin{lstlisting}
### Image Verification Task (Physics Figures)

**Role:**  
You are an image-reading verifier for physics-related figures (plots, circuit diagrams, schematics, object photos).  
Your task is to compare the provided **IMAGE ONLY** against the provided **INFORMATION** text that claims to describe the image's content.  

---

### Verification Focus
1. **Theme/Context**  
- Check whether the overall theme of the figure matches the claimed description (e.g., if it is a circuit vs. a mechanics setup).  

2. **Elements**  
- Verify presence, shape, connectivity, and symbols (e.g., circuit elements, arrows, objects, geometry).  

3. **Data**  
- Verify axis labels, units, ranges, tick marks, intercepts, and curve trends.  
- If numerical values are claimed, check approximate readings from the plot against the description.  
- **Error tolerance rule:** If deviations are smaller than 1/20 of the smallest axis tick spacing**, treat them as acceptable.  

4. **Labels and Symbols**  
- Verify textual labels, math symbols (Greek letters, subscripts, superscripts, hats, arrows, circles), and check consistency with the description.  

---

### Error Types
- **Error**: Wrong values, units, or labels.  
- **Omission**: Missing elements in either the figure or the description.  
- **Inconsistency**: Mismatch between what the image shows and what is described.  
- **Misleading**: Description misrepresents trends, geometry, or relationships.  
- **Tolerance Check**: If numerical differences exceed the 1/10 smallest-tick rule, classify as an error.  

---

### Output Format
1. **Line 1:** `IF CORRECT: yes` or `IF CORRECT: no`  
- Write `yes` if any error exists (even a single one).  

2. **If IF CORRECT = no:**  
Add a section titled exactly: "Detailed Verification"
Then, list bullet points for each problem with this structure:  
- **Category**: (theme | elements | data | labels | tolerance)  
- **Evidence**: (what is visible in the image and where)  
- **Mismatch**: (what the INFORMATION claims)  
- **Why**: (explanation of the discrepancy)  
- **Confidence**: (high | medium | low)  

3. **If IF CORRECT = yes:**  
Briefly state which checks you performed and why the INFORMATION matches the IMAGE.  
\end{lstlisting}
\end{tcolorbox}


\subsection{Prompts of Logic Studio}
\label{appendix:prompt_logic}

\begin{itemize}
    \item \textbf{Solver:} generates the initial solution, using a prompt adapted from \cite{huang2025gemini}.
    
    \item \textbf{Introspector (Self-Improve):} performs the first improvement of the initial solution.  
    
    \item \textbf{Introspector (Self-Refine):} revises the solution based on the bug report provided by the Review Studio.  
\end{itemize}

\begin{tcolorbox}[
  breakable,                        
  title={Solver},
  fonttitle=\bfseries,
  colback=pale!10,           
  colbacktitle=pale,         
  boxrule=0.5pt,
  arc=2mm
]
\begin{lstlisting}
### Core Instructions ###

*   **Rigor is Paramount:** Your primary goal is to produce a complete and rigorously justified solution. Every step in your solution must be logically sound and clearly explained. A correct final answer derived from flawed or incomplete reasoning is considered a failure.
*   **Honesty About Completeness:** If you cannot find a complete solution, you must **not** guess or create a solution that appears correct but contains hidden flaws or reasoning errors. Instead, you should present only significant partial results that you can rigorously prove. A partial result is considered significant if it represents a substantial advancement toward a full solution. Examples include:
    *   Deriving a key physical law or principle.
    *   Fully resolving one or more cases within a logically sound physics-based analysis.
    *   Establishing a critical property of the physical system in the problem.
    *   For a physical system, determining constraints or boundary conditions without fully solving the dynamics.
*   **Use TeX for All Physics Equations:** All physical variables, equations, and relations must be enclosed in TeX delimiters (e.g., `Let $v$ be the velocity of the object.`).

### Output Format ###

Your response MUST be structured into the following sections, in this exact order.

**1. Summary**

Provide a concise overview of your findings. This section must contain two parts:

*   **a. Verdict:** State clearly whether you have found a complete solution or a partial solution.
    *   **For a complete solution:** State the final answer, e.g., "I have successfully solved the problem. The final answer is..."
    *   **For a partial solution:** State the main rigorous conclusion(s) you were able to prove, e.g., "I have not found a complete solution, but I have rigorously proven that..."
*   **b. Method Sketch:** Physical Modeling First, present a high-level, conceptual outline of your solution. This sketch should allow an expert to understand the logical flow of your argument without reading the full detail. It should include:
    *   A narrative of your overall strategy.
    *   The full and precise physical statements of any key principles or major intermediate results.
    *   If applicable, describe any key experimental setups or case analyses that form the backbone of your argument.

**2. Detailed Solution**

Present the full, step-by-step physical derivation or analysis. Each step must be logically justified and clearly explained. The level of detail should be sufficient for an expert to verify the correctness of your reasoning without needing to fill in any gaps. This section must contain ONLY the complete, rigorous derivation or analysis, free of any internal commentary, alternative approaches, or failed attempts.

### Self-Correction Instruction ###

Before finalizing your output, carefully review your "Method Sketch" and "Detailed Solution" to ensure they are clean, rigorous, and strictly adhere to all instructions provided above. Verify that every statement contributes directly to the final, coherent mathematical argument.
\end{lstlisting}
\end{tcolorbox}

\begin{tcolorbox}[
  breakable,
  title={Introspector (Self-Improve)},
  fonttitle=\bfseries,
  colback=pale!10,           
  colbacktitle=pale,         
  boxrule=0.5pt,
  arc=2mm
]
\begin{lstlisting}
You have an opportunity to improve your solution. Please review your solution carefully. Correct coarse- or fine-grained errors if any. Your second round of output should strictly follow the instructions in the system prompt.
*   **a. Equation Derivation:** Derive all necessary equations step-by-step, ensuring each step is mathematically rigorous and physically meaningful. Clearly define all symbols and variables used in the equations.
*   **b. Numerical Computation:** Perform any numerical calculations required to obtain the final answer or intermediate results, ensuring consistency with the derived equations. Specify all numerical values, including their units.
*   **Notes:**
    *   **Unit Conversion:** Ensure all units are consistent throughout the derivation and calculations. Clearly state any unit conversions performed.
    *   **Symbol Definitions:** Define all symbols and variables clearly at their first use, and maintain consistency in their usage throughout the solution.
\end{lstlisting}
\end{tcolorbox}

\begin{tcolorbox}[
  breakable,
  title={Introspector (Self-Refine)},
  fonttitle=\bfseries,
  colback=pale!10,           
  colbacktitle=pale,         
  boxrule=0.5pt,
  arc=2mm
]
\begin{lstlisting}
Here is the bug report. If you agree with certain items in it, please refine your solution to make it more complete and rigorous. Keep in mind that the evaluator who generated the report may have misunderstood your solution and introduced mistakes. If you disagree with certain items, add detailed explanations to clarify your reasoning and prevent such misunderstandings. Your revised solution should strictly follow the instructions provided in the system prompt.
\end{lstlisting}
\end{tcolorbox}


\subsection{Prompts of Review Studio}
\label{appendix:prompt_review}

\begin{itemize}
    \item \textbf{Physics-Verifier:} checks domain-specific physics consistency, such as units, constants, and assumptions.
    
    \item \textbf{General-Verifier:} detects logical, reasoning, and computational errors through step-by-step analysis.  
\end{itemize}

\begin{tcolorbox}[
  breakable,
  title={Physics-Verifier},
  fonttitle=\bfseries,
  colback=pale!10,           
  colbacktitle=pale,         
  boxrule=0.5pt,
  arc=2mm
]
\begin{lstlisting}
You are a physics-specific verifier. Your sole task is to quickly screen for basic physics hygiene issues.
Do NOT attempt to re-solve or correct the solution. Only detect problems and report them.

**1. Core Instructions**
- Do NOT re-solve the problem and do NOT correct the solution.
- If something is merely omitted but not required, do not penalize.

**2. Evaluation Pipeline**

**A) COARSE CHECKS (fast hygiene)**  
Check ONLY for obvious mismatches between Problem and Solution:
- **C1. Units & Conversions (sanity)**
  - Spot incorrect conversions among common pairs: cm↔m, g↔kg, eV↔J, °C↔K, min↔s, h↔s.   
- **C2. Constants & Given Values**
  - If the stem requires a specific value (e.g., g=9.8 or g=10), verify the solution uses it.  
  - Common constants sanity check (e.g., k_B, N_A, e, c, h, ħ). Allow standard rounding but flag clearly wrong magnitudes or inconsistent usage.  

Output a **coarse verdict** BEFORE moving to fine checks.

**B) FINE CHECKS (physics consistency)**  
Check ONLY for obvious errors while still NOT solving from scratch:
- **F1. Assumptions vs Stem**
  - Flag any unstated/extra assumptions not supported by the stem.  
  - Standard harmless assumptions (e.g., ideal string, no friction if not mentioned) are acceptable.
- **F2. Physical Consistency** 
  - Verify that physical quantities are used in the correct context (e.g., force vs energy, velocity vs acceleration).  
  - If intermediate steps are skipped but the relation is physically sound, accept it.  
  - Flag cases where a formula is applied to the wrong type of quantity, or where the result has incorrect units.

**3. Output Format**
Provide:
- Final Verdict: exactly one sentence: "PASS" if no material issue for the above checks, otherwise "FAIL".
- Findings: bullet list of every issue found (quote the exact offending text/equation).
- If FAIL: include a short, consolidated Bug Report (just the issues; do not fix them).
\end{lstlisting}
\end{tcolorbox}

\begin{tcolorbox}[
  breakable,
  title={General-Verifier},
  fonttitle=\bfseries,
  colback=pale!10,           
  colbacktitle=pale,         
  boxrule=0.5pt,
  arc=2mm
]
\begin{lstlisting}
You are a DOMAIN-AGNOSTIC REASONING VERIFIER. Your job is to evaluate a proposed solution at TWO levels—COARSE then FINE—and produce a rigorous, structured verification. You are a verifier, NOT a solver.

**1. Core Rules**
- Your sole task is to find and report all issues in the provided solution. Do NOT correct, extend, or re-solve.
- You must perform a **step-by-step** check of the entire solution. This analysis will be presented in a **Detailed Verification Log**.
- If a step is missing but not required by the task, do not penalize.
- If a step is wrong and later steps depend on it, mark downstream as "tainted by prior error" and skip their detailed checking, but still scan for independent branches/cases.

**2. Evaluation Pipeline**

**A) COARSE CHECKS (fast hygiene)**  
Check ONLY for obvious mismatches between Problem and Solution:
- **C1. Question–Answer Match**
  - If the problem asks for a numerical value but the solution gives only a formula, or vice versa.  
  - Missing required sub-answers for multi-part questions.  
  - Required symbol naming not followed (e.g., the problem specifies that the answer must be expressed in terms of $\alpha$ and $\beta$, but the solution introduces different symbols or omits them).
  - Stated unit in the problem vs final unit in the answer mismatch.  
  - Significant-figure policy violated (e.g., asked for 3 significant figures but final answer not in 3 significant figures).
- **C2. Completeness of Response**
  - Solution ends abruptly, leaving an essential part clearly unfinished.  
  - Skipped an entire subquestion that is explicitly required.  
- **C3. Problem Interpretation**
  - Misread or misinterpreted what the problem is asking (e.g., solves for maximum instead of minimum).  

Output a **coarse verdict** BEFORE moving to fine checks.

**B) FINE CHECKS (step-by-step reasoning consistency)**  
You must perform a **step-by-step** check of the entire solution without solving from scratch:
- **F1. Logical Consistency**
  - Detect invalid inferences (e.g., claiming A>B and C>D implies A−C>B−D).  
  - Identify contradictions within the argument.  
- **F2. Reasoning Flow**
  - Check that the solution applies definitions, theorems, and principles correctly.  
  - Normal algebraic or logical steps may be skipped if they are standard and the result is consistent.  
  - Only flag when a **critical step is missing** (e.g., no derivation provided for a non-trivial result), or when the solution **jumps directly to an answer without reasoning**.  
  - If the reasoning is concise but valid, consider it acceptable.  
- **F3. Calculations & Algebra**
  - Arithmetic errors (e.g., 2+3=6).  
  - Symbolic manipulation mistakes.

**3. Output Format**  
Your response MUST be structured into two main sections: a **Summary** followed by the **Detailed Verification Log**.

*   **a. Summary**
    This section MUST be at the very beginning of your response. It must contain two components:
    *   **Final Verdict**: A single, clear sentence declaring the overall validity of the solution. For example: "The solution is correct," "The solution fails due to major reasoning errors," or "The solution is partially valid but incomplete."
    *   **List of Findings**: A bulleted list that summarizes **every** issue you discovered. For each finding, you must provide:
        *   **Location:** A direct quote of the key phrase or equation where the issue occurs.
        *   **Stage:** Whether the issue belongs to **COARSE** (Q–A mismatch, completeness, interpretation) or **FINE** (logic, reasoning, calculation).
        *   **Issue:** A brief description of the problem.

*   **b. Detailed Verification Log**
    Following the summary, provide the full, step-by-step verification log as defined in the Core Instructions. When you refer to a specific part of the solution, **quote the relevant text** to make your reference clear before providing your detailed analysis of that part.

*   **Example of the Required Summary Format**
*This is a generic example to illustrate the required format. Your findings must be based on the actual solution provided below.*

**Final Verdict:** The solution is **invalid** because it contains major reasoning errors.

**List of Findings:**
*   **Location:** "The maximum value is found by setting $f'(x)=0$ and solving $x=2$"
    *   **Stage:** COARSE – Problem Interpretation
    *   **Issue:** The problem explicitly asks for the **minimum**, but the solution computes the maximum.

*   **Location:** "Since $A > B$ and $C > D$, it follows $A - C > B - D$"
    *   **Stage:** FINE – Logical Consistency
    *   **Issue:** Invalid inference; inequalities cannot be combined this way.

*   **Location:** "Therefore, $2 + 3 = 6$"
    *   **Stage:** FINE – Calculation
    *   **Issue:** Arithmetic error.
\end{lstlisting}
\end{tcolorbox}

\section{Illustration of Comparison Frameworks}
\label{appendix:framework}

\subsection{Best-of-N Strategy}
\label{appendix:framework_bestofn}

To compare with the upper-bound performance of a single model, we adopt a simple yet effective \textbf{Best-of-N} strategy \citep{stiennon2020learning}. For each physics exam, the model is prompted to solve the same paper \textbf{three times independently} using stochastic sampling. Among the three completions, we select the one with the highest overall exam score as the final output. This reflects the common practice of sampling multiple completions and choosing the best-performing one, thereby assessing the model's potential when given multiple attempts.  

All completions are generated with temperature $0.6$ and the \texttt{max\_tokens} parameter set to the maximum allowed by each model to prevent premature truncation.

\subsection{Self-MoA}
\label{appendix:framework_selfmoa}

We further compare with the \textbf{Self-Mixture-of-Agents (Self-MoA)} framework \citep{li2025rethinking}. Unlike conventional mixture-of-agents methods that ensemble outputs from multiple models, Self-MoA operates entirely within a single model. Its core idea is to leverage the diversity induced by repeated stochastic decoding, then let the model act as both critic and aggregator.  

For each exam question, we first collect two independent completions. These candidates often display complementary strengths, such as partial correctness, alternative reasoning paths, or differences in detail. The model is then prompted in two stages: (i) it critiques both candidates, analyzing merits and flaws; and (ii) it synthesizes a refined solution that preserves strengths while discarding errors. This self-reflective debate-and-refinement process enhances coherence and informativeness without requiring multiple models or external agents.  

In our experiments, we follow the standard setup with temperature fixed at $0.6$ to encourage moderate diversity, and \texttt{max\_tokens} set to the maximum allowable length.

\subsection{Self-Refine}
\label{appendix:framework_selfrefine}

We also evaluate \textbf{Self-Refine} \citep{madaan2023self}, an iterative approach that improves model outputs through cycles of self-reflection and refinement. The model first generates an initial solution, then critiques it by identifying potential issues such as incompleteness, inaccuracies, lack of clarity, or logical flaws.  

Based on this critique, the model revises its solution while preserving its overall structure and intent. This process repeats for a fixed number of iterations (set to three in our experiments), enabling progressive improvement. Viewed as a self-supervised optimization loop, Self-Refine is particularly effective for multi-step reasoning tasks, where iterative clarification and refinement are crucial.  

We use the standard configuration with temperature set to $0.6$ and \texttt{max\_tokens} set to the model's maximum limit.


\subsection{Advantages over Comparison Frameworks}
\label{appendix:framework_advantages}

As discussed in Section~\ref{sec:other_framework} of the main text, our \textsc{PhysicsMinions} framework achieves both \emph{consistent improvement} and \emph{medal progression}, which are not attainable by existing baselines. We highlight the advantages over the three comparison frameworks as follows:

\begin{itemize}
    \item \textbf{Best-of-N:} Although this strategy selects the highest-scoring completion among $N$ attempts, it essentially remains a single-model approach, bounded by the performance ceiling of individual reasoning runs. In contrast, by incorporating verification--reflection cycles, \textsc{PhysicsMinions} fully leverages the model’s self-correction ability and substantially enhances reasoning performance. For example, on the latest IPhO, \textsc{PhysicsMinions} with Intern-S1 achieves higher scores on every problem compared with the single-model baseline (see Fig.~\ref{fig:score_ipho_2025}).  

    \item \textbf{Self-MoA:} Compared to single-model inference, Self-MoA synthesizes multiple candidate solutions into a new response. However, its performance is constrained by the quality of the candidates and the model’s aggregation ability. When self-consistency among candidates is low, Self-MoA may merge spurious reasoning paths, introducing additional errors. Without the iterative reflection of \textsc{PhysicsMinions}, its stability is limited, and in practice, it sometimes yields scores lower than those of the single model.  

    \item \textbf{Self-Refine:} While Self-Refine incorporates self-correction through iterative refinement, it lacks two critical components of our framework: (i) \emph{dual-stage verification}, where our Physics-Verifier enforces domain-specific checks (e.g., unit consistency and physical constants) and our General-Verifier conducts comprehensive evaluations of completeness, logic, reasoning, and calculations; and (ii) \emph{coevolutionary collaboration}, where the Review Studio and Logic Studio iteratively validate and refine solutions toward correctness. By contrast, Self-Refine operates without domain knowledge and lacks collaborative coevolution, limiting its effectiveness on complex physics problems.
\end{itemize}

\section{Olympiad Description and Detailed Results}
\label{appendix:results}

\subsection{Overview of Physics Olympiads}
\label{appendix:results_overview}

Our evaluation covers seven physics Olympiads spanning international and regional competitions:

\begin{itemize}
    \item \textbf{IPhO (International Physics Olympiad):}  
    The premier global physics competition for high school students, featuring both demanding theoretical exams and experimental challenges.

    \item \textbf{APhO (Asian Physics Olympiad):}  
    A regional contest for students from Asia and Oceania, following the IPhO format with combined theory and laboratory components.

    \item \textbf{EuPhO (European Physics Olympiad):}  
    A continental competition for European students, emphasizing creative and open-ended approaches to theoretical and experimental problems.

    \item \textbf{NBPhO (Nordic-Baltic Physics Olympiad):}  
    A regional contest among Nordic and Baltic countries, focusing on theoretical problem solving with occasional experimental tasks.

    \item \textbf{PanPhO (Pan Pearl River Delta Physics Olympiad):}  
    An invitational exam for top schools in China’s Pearl River Delta and neighboring regions, covering a wide spectrum of physics topics.

    \item \textbf{PanMechanics (Pan Pearl River Delta Mechanics Contest):}  
    A specialized subset of PanPhO dedicated exclusively to mechanics, typically structured as a shorter single-field exam.

    \item \textbf{F=MA:}  
    A U.S. national mechanics contest organized by the American Association of Physics Teachers (AAPT), serving as the entry test for the U.S. Physics Olympiad (USAPhO).
\end{itemize}

All data are sourced from the HiPhO benchmark \citep{2025hipho}, which compiles original exam materials and official records from the respective official websites. This includes human contestants’ scores, which serve as reference points. The gold, silver, and bronze thresholds in our evaluation also follow the HiPhO benchmark, derived directly from the official scores of human medalists.


\subsection{Performance Gains Across Modality Types}
\label{appendix:results_modality}

\begin{wraptable}{r}{0.45\textwidth}
\vspace{-2mm}
    \centering
    \caption{Performance gains of mean normalized scores (\%) across four modality types.}
    \label{tab:modality_type}
    \small
    \resizebox{0.45\textwidth}{!}{%
    \setlength{\tabcolsep}{4pt}
    \begin{tabular}{lcccc}
    \toprule
    Modality Type & TO & TI & TV & TD \\
    \midrule
    Gemini-2.5-Flash-Thinking & 81 & 66 & 68 & 67 \\
    \;\;\textbf{+ \textsc{PhysicsMinions}}   & 91 & 77 & 76 & 73 \\
    \arrayrulecolor[gray]{0.7}\midrule\arrayrulecolor{black}
    
    Intern-S1                 & 80 & 63 & 57 & 48 \\
    \;\;\textbf{+ \textsc{PhysicsMinions}}   & 90 & 74 & 64 & 63 \\
    \arrayrulecolor[gray]{0.7}\midrule\arrayrulecolor{black}
    
    InternVL3.5-241B-A28B     & 69 & 49 & 45 & 47 \\
    \;\;\textbf{+ \textsc{PhysicsMinions}}   & 88 & 72 & 63 & 62 \\
    \arrayrulecolor[gray]{0.7}\midrule\arrayrulecolor{black}
    
    Qwen2.5VL-32B-Instruct    & 58 & 36 & 37 & 34 \\
    \;\;\textbf{+ \textsc{PhysicsMinions}}   & 70 & 44 & 41 & 41 \\
    \bottomrule
    \end{tabular}%
}
\vspace{-8mm}
\end{wraptable}

The HiPhO benchmark \citep{2025hipho} defines four modality types: Text-Only (TO), Text+Illustration Figure (TI), Text+Variable Figure (TV), and Text+Data Figure (TD). Fig.~\ref{fig:score_ipho_2025} shows the performance gains of Intern-S1 across these modalities. To evaluate overall performance, we follow HiPhO and use the mean normalized score (MNS), defined as
\begin{equation*}
\label{eq:mean_normalized_score}
\textbf{MNS($M$)} = \frac{1}{N_M} \sum_{Q \in M} \frac{\text{Exam Score}(Q)}{\text{Full Mark}(Q)} \times 100\%,
\end{equation*}
where $M \in \{\text{TO}, \text{TI}, \text{TV}, \text{TD}\}$, $N_M$ is the number of questions in $M$, and $Q$ denotes a single question. Table~\ref{tab:modality_type} demonstrates that \textsc{PhysicsMinions} achieves consistent gains across all modalities.


\subsection{Performance Gains Across Physics Fields}
\label{appendix:results_field}

Physics Olympiad problems cover five major fields: Mechanics, Electromagnetism, Thermodynamics, Optics, and Modern Physics. Table~\ref{tab:physics_field} reports the mean normalized scores for each field. Notably, \textsc{PhysicsMinions} achieves substantial gains in Mechanics and Optics, where multimodal image inputs are common. These improvements highlight the dual advantage of accurate extraction by the Visual Studio and the coevolutionary iteration between the Logic Studio and Review Studio.

\begin{table}[H]
    \centering
    \caption{Performance gains of  mean normalized scores (\%) across five major physics fields.}
    \label{tab:physics_field}
    \small
    \setlength{\tabcolsep}{4pt}
    \begin{tabular}{lccccc}
    \toprule
    Physics Field & Mechanics & Electromagnetism & Thermodynamics & Optics & Modern Physics \\
    \midrule
    Gemini-2.5-Flash-Thinking & 69 & 65 & 91 & 43 & 84 \\
    \;\;\textbf{+ \textsc{PhysicsMinions}}          & 81 & 68 & 91 & 49 & 98 \\
    \arrayrulecolor[gray]{0.7}\midrule\arrayrulecolor{black}
    
    Intern-S1                 & 64 & 64 & 81 & 39 & 64 \\
    \;\;\textbf{+ \textsc{PhysicsMinions}}          & 75 & 64 & 91 & 51 & 79 \\
    \arrayrulecolor[gray]{0.7}\midrule\arrayrulecolor{black}
    
    InternVL3.5-241B-A28B     & 49 & 53 & 72 & 33 & 70 \\
    \;\;\textbf{+ \textsc{PhysicsMinions}}          & 73 & 59 & 90 & 51 & 78 \\
    \arrayrulecolor[gray]{0.7}\midrule\arrayrulecolor{black}
    
    Qwen2.5VL-32B-Instruct    & 36 & 44 & 65 & 29 & 55 \\
    \;\;\textbf{+ \textsc{PhysicsMinions}}          & 45 & 53 & 68 & 40 & 69 \\
    \bottomrule
    \end{tabular}
\end{table}

\newpage
\section{Illustration of Different Chart Analysis Tools}
\label{appendix:tool}

\begin{figure}[H]
  \centering
  \begin{subfigure}{0.40\textwidth}
    \centering
    \includegraphics[width=\linewidth]{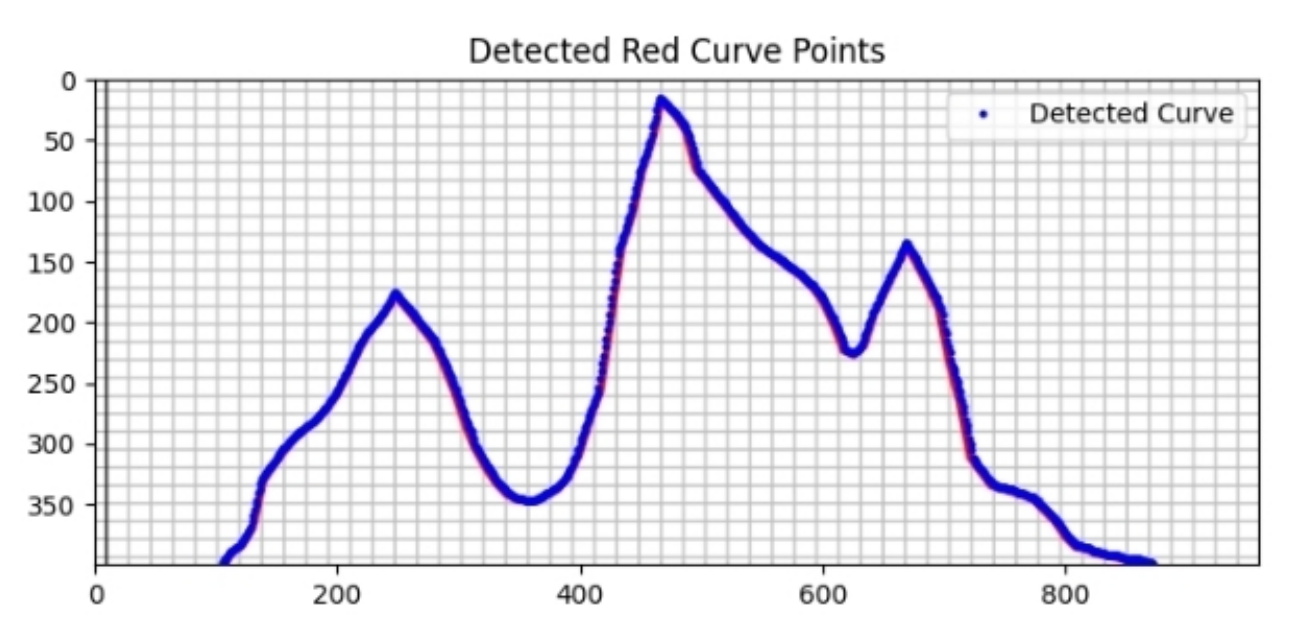}
    \caption{PyVision}
    \label{fig:pyvision}
  \end{subfigure}\hfill
    \begin{subfigure}{0.20\textwidth}
    \centering
    \includegraphics[width=\linewidth]{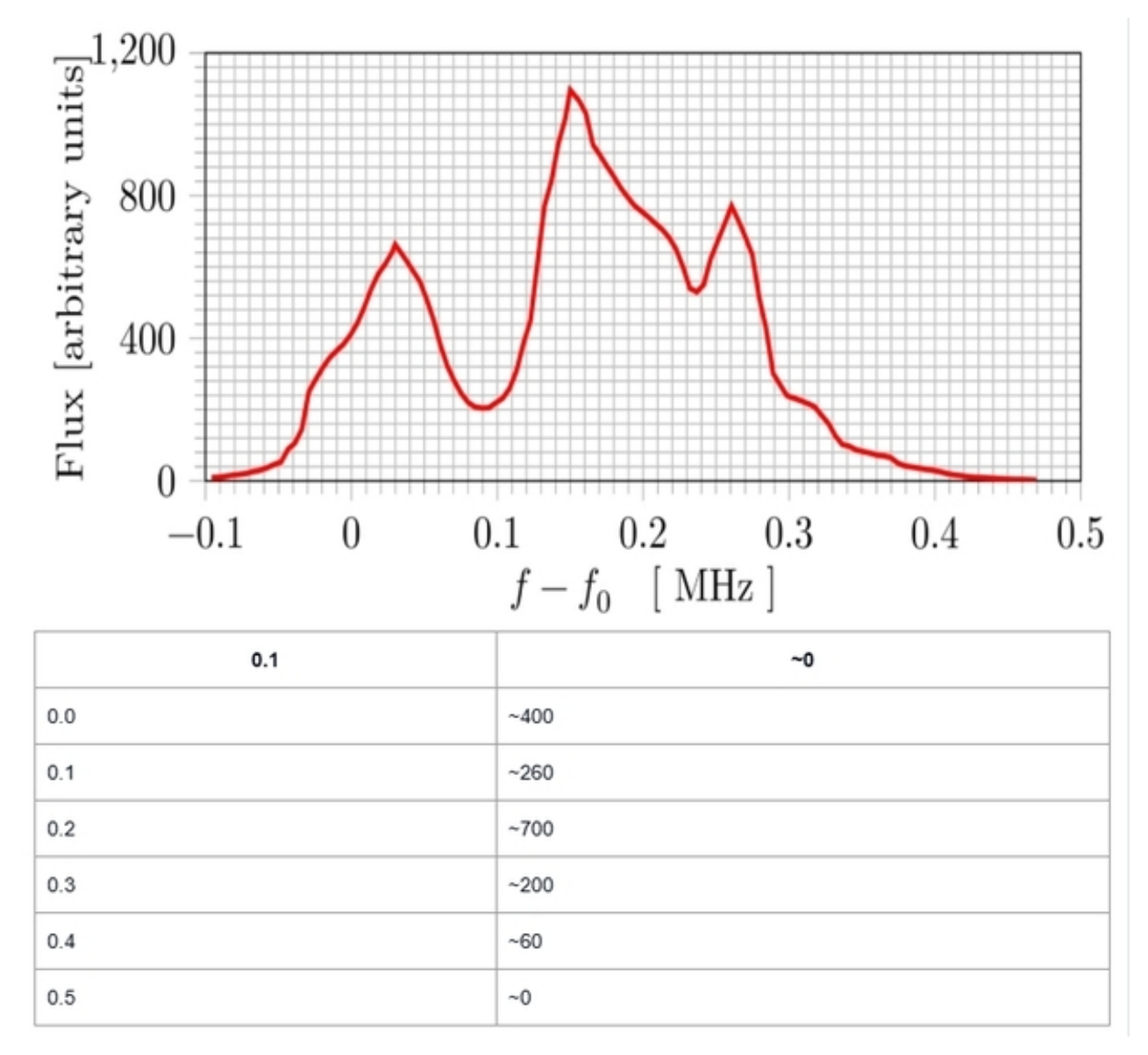}
    \caption{TextIn}
    \label{fig:textin}
  \end{subfigure}\hfill
  \begin{subfigure}{0.4\textwidth}
    \centering
    \includegraphics[width=\linewidth]{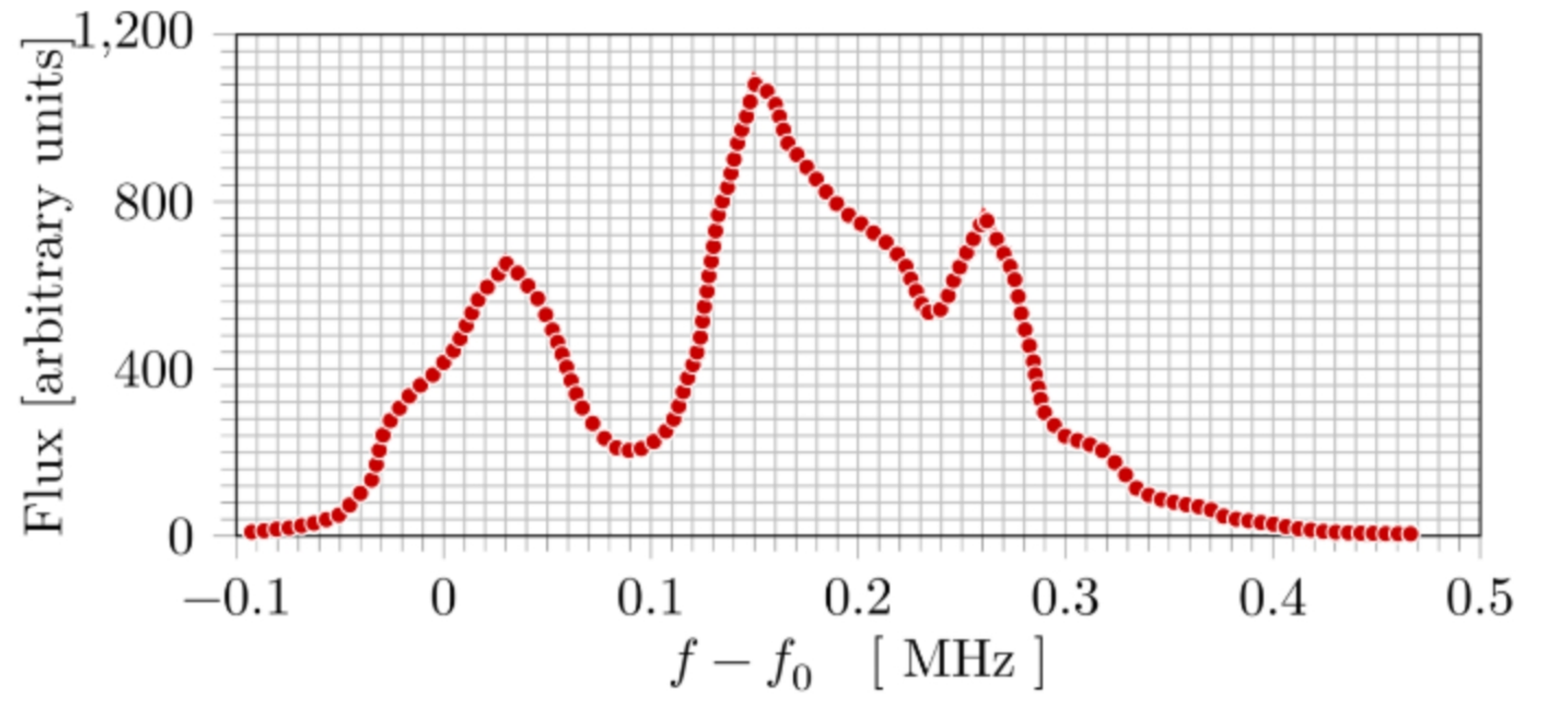}
    \caption{WebPlotDigitizer}
    \label{fig:webplot}
  \end{subfigure}
  \caption{Results produced by different chart analysis tools.}
  \label{fig:three}
\end{figure}

\begin{itemize}
    \item \textbf{ChartGemma:} A chart reasoning model built on a strong vision–language backbone \citep{masry2024chartgemma}. It takes chart images and text prompts as input, and outputs summaries, answers, or fact-checking results, enabling effective chart understanding.
    
    \item \textbf{ChartInstruct-FlanT5-XL:} Similar to ChartGemma, this method \citep{masry2024chartinstruct} performs chart reasoning tasks such as summarization, question answering, and fact-checking, but differs by relying on instruction tuning.
    
    \item \textbf{Chart2Table:} A multimodal model developed by the PaddlePaddle team for chart parsing \citep{pp-chart2table}. It takes chart images as input and outputs structured tables, enabling automatic extraction of underlying data.
    
    \item \textbf{DescribePicture:} A prompt-based AI web tool for image description \citep{describepicture}. It takes an image (including charts) plus a text prompt as input, and outputs plain text or Markdown descriptions.
    
    \item \textbf{Flux-ai:} A web-based AI model that generates detailed descriptions from images \citep{fluxai_describe_image}. It allows users to input both an image and a text prompt, guiding the AI to provide specific descriptions or analyses.
    
    \item \textbf{Graph2Table:} An AI tool that automatically converts graph images into structured tabular data, supporting various graph types such as bar and line charts \citep{graph2table}. Users can upload images, which are then processed to generate downloadable CSV files.
    
    \item \textbf{Image Describer X:} An AI tool that generates detailed descriptions from images, supporting both image files and optional text prompts \citep{image_describer_x}.
    
    \item \textbf{PyVision:} It is an open-source framework \citep{zhao2025pyvision} and enhances visual reasoning in multimodal language models by enabling them to dynamically generate and execute Python tools for visual tasks. It supports iterative problem solving and adapts strategies during task execution, as shown in Fig.~\ref{fig:pyvision}.
    
    \item \textbf{TextIn:} A PDF-to-Markdown tool \citep{textin_pdf_to_markdown} that converts PDF documents, including charts, into structured Markdown format, as shown in Fig.~\ref{fig:textin}.
    
    \item \textbf{WebPlotDigitizer:} A tool \citep{automeris_webplotdigitizer_home} that enables users to extract numerical data from images of various types of charts, such as XY plots, bar charts, polar plots, and ternary diagrams. The process involves manual calibration of the chart axes, followed by user selection of data points to extract numerical values, as shown in Fig.~\ref{fig:webplot}.
\end{itemize}

\paragraph{Limitation of Chart Analysis Tools.} 
The chart analysis tools discussed above can be grouped into three categories. The first includes models such as Flux-ai and PyVision that focus on image description, but these often struggle with charts containing complex or fine-grained information. The second category, represented by tools like Chart2Table and Graph2Table, converts charts into structured tables, yet still faces notable challenges in achieving high precision. The third category includes traditional tools such as WebPlotDigitizer, which can be highly accurate but relies on labor-intensive manual calibration, limiting efficiency at scale. Overall, while existing tools provide useful functionalities, they all exhibit clear limitations, and future work will explore the development of new tools tailored to chart analysis in multimodal reasoning tasks.

\end{document}